\documentclass{article}

\usepackage[preprint]{neurips_2021}

\usepackage[utf8]{inputenc} % allow utf-8 input
\usepackage[T1]{fontenc}    % use 8-bit T1 fonts
\usepackage[hyphens]{url}
\usepackage{hyperref}       % hyperlinks
\usepackage{url}            % simple URL typesetting
\usepackage{booktabs}       % professional-quality tables
\usepackage{amsfonts}       % blackboard math symbols
\usepackage{nicefrac}       % compact symbols for 1/2, etc.
\usepackage{microtype}      % microtypography
\usepackage{xcolor}         % colors
\usepackage{graphicx}
\usepackage{tikz}
\usepackage{pgfplots}
\usepackage{adjustbox}
\usepackage{subfigure}
\usepackage{xcolor}
\usepackage{colortbl}
\usepackage{wrapfig}
\usepackage{algorithm,algorithmicx,algpseudocode}
\usepackage[T1]{fontenc}
\usepackage[utf8]{inputenc}
\usepackage{babel}
\usepackage[font=small,labelfont=bf]{caption}
\usepackage{multirow}
\usepackage{listings}
\usepackage{soul}
\title{Logit Attenuating Weight Normalization}
\author{%
  Aman Gupta\thanks{LinkedIn Corporation. (\texttt{@linkedin.com})} \\
  \small{\texttt{amagupta}}
  \And
  Rohan Ramanath\footnotemark[1]\\
  \small{\texttt{rramanath}}
  \And
  Jun Shi\footnotemark[1] \\
  \small{\texttt{jshi}}
  \AND
  Anika Ramachandran\thanks{University of California, Berkeley (\texttt{@berkeley.edu})} \\
  \small{\texttt{anikar}}
  \And 
  Sirou Zhou\thanks{Carnegie Mellon University (\texttt{@andrew.cmu.edu})} \\
  \small{\texttt{sirouz}}
  \And
  Mingzhou Zhou\footnotemark[1]\\
  \small{\texttt{mizhou}}
  \And
  S. Sathiya Keerthi\footnotemark[1] \\
  \small{\texttt{keselvaraj}} \\
}
\begin{document}

\maketitle

\vspace{-0.5cm}

\begin{abstract}
Over-parameterized deep networks trained using gradient-based optimizers are a popular choice for solving classification and ranking problems. Without appropriately tuned $\ell_2$ regularization or weight decay, such networks have the tendency to make output scores (logits) and network weights large, causing training loss to become too small and the network to lose its adaptivity (ability to move around) in the parameter space. Although regularization is typically understood from an overfitting perspective, we highlight its role in making the network more adaptive and enabling it to escape more easily from weights that generalize poorly. To provide such a capability, we propose a method called Logit Attenuating Weight Normalization (LAWN), that can be stacked onto any gradient-based optimizer.  LAWN controls the logits by constraining the weight norms of layers in the final homogeneous sub-network. Empirically, we show that the resulting LAWN variant of the optimizer makes a deep network more adaptive to finding minimas with superior generalization performance on large scale image classification and recommender systems. While LAWN is particularly impressive in improving Adam, it greatly improves all optimizers when used with large batch sizes.
\end{abstract}

\vspace{-0.75cm}

\section{Introduction}
\label{sec:intro}

%\keerthi{This is a placeholder introduction. It will be rewritten during this week. It is also worth mentioning that adding L2 for deep nets is not to avoid overfitting, but to avoid loss flattening. Also discuss the relationship b/w L2 constant and batch size.}

%The study of gradient-based optimization methods has resulted in the development of optimizers like Stochastic Gradient Descent (SGD) with heavy-ball momentum, Adam~\citep{kingma2017adam}, AdamW~\citep{loshchilov2019decoupled}, RAdam, AdaHessian and their extensions with $\ell_2$ regularization and weight decay to improve generalization performance.%

The advent of large scale deep models with tens of millions to billions of parameters has resulted in three trends in the community: (1) State-of-the-art performance on over-parameterized networks for problems like image classification, language modeling, machine translation, text classification and recommender systems. (2) Development of optimizers like Stochastic Gradient Descent (SGD) with heavy-ball momentum~\citep{qian1999momentum}, Adam~\citep{kingma2017adam}, AdamW~\citep{loshchilov2019decoupled}, LAMB~\citep{you2020large} and their extensions with $\ell_2$ regularization and weight decay to improve generalization performance. (3) Increased emphasis on theory to understand the optimizer landscape,  especially how to tune different hyperparameters well to escape poor minima for both large and small batch sizes. In this work, inspired by all three trends, we propose a new training method, explain why it works, and show improved performance over popular optimizers across a wide range of batch sizes for classification and ranking tasks. %using over-parameterized networks.
% increased emphasis on understanding the geometric properties of regions around minima and the role played by hyperparameters in going to more generalizable minima; for both small and large batch sizes. the increased emphasis on theory to understand the path taken by each optimizer and geometric properties of the region around the minima has helped tune different hyperparameters and learning rate schedules to escape poor minima even at large batch sizes.

%\keerthi{Rohan to re-write this, and wherever this kind of text appears in the paper. Talk about gradients becoming small even when 80-90\% of training samples become "correctly" classified. This leads to large weights nonetheless, lower loss values and subsequently lower gradients. This makes the network less adaptive. Future work - what if majority of examples are noisy?}

Complex deep networks can easily learn to classify a large fraction of examples correctly 
%(e.g. $80-100\%$ depending on noise) 
as training progresses. %since they can always memorize the training dataset in the absence of noise. 
These networks have two characteristics: (a) they have one or more contiguous \textit{homogeneous\footnote{A layer is {\em homogeneous} if the activation function of each unit of the layer  satisfies $\phi(\alpha x) = \alpha\phi(x)$. Linear and ReLU are examples of such activation functions.}} layers at the end forming the final homogeneous sub-network, and, (b) they are trained with exponential-type loss functions like logistic loss and cross entropy that asymptotically attain their least value of zero when the network score goes to infinity. Let us collectively refer to this network score as {\em logit}. After the network has learned to correctly classify a large fraction of training examples,  the weights of the end homogeneous layers and the logits grow to make the training loss (and hence its gradient) very small. This, seen in optimizers like SGD~\citep{Bottou2010} and Adam ~\citep{kingma2017adam} when used with no (or mild) $\ell_2$ regularization or weight decay, leads to {\em loss flattening} (Details in Appendix \ref{sec:lossflattening}). This further leads to {\em loss of adaptivity} of the network~\citep{szegedy2016rethinking}, causing training to stall in regions of sub-optimal generalization. %In other words, the large majority of correctly classified examples make it harder for the network to explore a different region of the parameter space that might help it correctly classify the few incorrectly labeled examples. \rohan{what do people think of the last line?}

%At the stage in training when the network can correctly classify most examples, the logits and  weights of the network start to grow which subsequently results in lower loss and hence gradients. This makes the network less adaptive to explore regions of the parameter space that will help it learn to classify the hard examples as well as generalize better on unseen data. 

%The combination of over-parameterized networks, gradient-based optimizers, and classification or ranking problems is useful because (a) over-parameterized multilayer neural nets can easily achieve perfect or near-perfect training accuracy, and (b) exponential type loss functions such like logistic loss and cross entropy asymptotically go to their least value of zero when the network score goes to infinity. 

%The network score is either the logit of the target class for binary classification or the relative score of the target class against other classes for multi-class classification. Additionally, all networks used for classification 

%When optimizers like Stochastic Gradient Descent (SGD) (cite) and Adam (cite) are trained without explicit regularization or weight decay, they increase the weights of the end homogeneous layers and make the training loss very small. This leads to {\em loss flattening} and hence {\em loss of adaptivity} of the network, causing the training to be caught in points of sub-optimal generalization. 
Figure~\ref{fig:lawn_motivation} shows the generalization performance (Test HR@10) of Adam (green dotted line) applied to a fully homogeneous network on a classification task. After $115$ epochs, {\em Margin p50} (median margin over the training examples; blue dotted line) becomes large and the training gets stuck in the basin of a sub-optimally generalizing minimum. The minimum is also overfitting due to the tussle between good and noisy examples.

%shows the performance of a fully homogeneous over-parameterized network on a classification task using Adam (dotted line) where the generalization performance drops after a point even through the training loss continues to decrease. We also show evidence (in blue) that gradient-based methods implicitly optimize the normalized margin as training progresses.

%\keerthi{This figure needs to be updated. Keerthi to add comment about the green drop. }
% \begin{figure}[h]
%     \centering
%     \includegraphics[width=0.6\columnwidth]{images/exp1-ml1mdemo-run1.png}
%     \vspace{-0.2cm}
%     \caption{Comparison of Adam and Adam-LAWN for a classification task. After around $110$ epochs, Adam starts overfitting because of {\em Margin p50} becoming very high. For Adam-LAWN, we shrink the network weights by a small positive factor and continue training while keeping the weight norms of each layer fixed.}
%     \label{fig:lawn_motivation}
% \vskip -0.1in
% \end{figure}
  \begin{minipage}{\textwidth}
  \begin{minipage}[b]{0.57\textwidth}
    \centering
    \includegraphics[width=1.0\columnwidth]{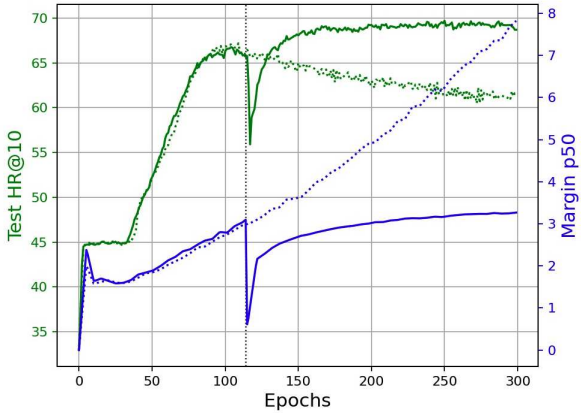}
    \label{fig:lawn_motivation}
    \vspace{-0.6cm}
    \captionof{figure}{Adam with (continuous lines) and without (dotted lines) logit attenuation on MovieLens classification. {\em Adam:} after $115$ epochs (vertical dotted black line), loss flattening sets in, Test HR@10 is sub-optimal, with overfitting. {\em Adam with logits attenuation:} The introduction of $\alpha$ factor after the $115$th epoch reduces {\em Margin p50} but keeps Test HR@10 the same. Though Test HR@10 then drops initially, with further constrained weight training Adam with logit attenuation eventually achieves a higher Test HR@10 than vanilla Adam.
    %With further constrained weight training, though Test HR drops initially, it turns around and reaches a much larger value than vanilla Adam at the end.%
    }
  \end{minipage}
  \hfill
  \begin{minipage}[b]{0.41\textwidth}
    \centering
\resizebox{0.95\textwidth}{!}{
\begin{tabular}{@{}lll@{}}
\toprule
%\rowcolor[HTML]{FFFFFF} 
\multirow{2}{*}{\textbf{Optimizer}} & \multicolumn{2}{c}{\textbf{Batch Size}} \\
 & \multicolumn{1}{c}{256} & \multicolumn{1}{c}{16k} \\ 
\midrule
%\rowcolor[HTML]{FFFFFF} 
SGD                              & 76.00 (0.04)           & 74.48 (0.06)            \\
%\rowcolor[HTML]{FFFFFF} 
SGD-L                         & \textbf{76.12} (0.01)           & \textbf{75.56} (0.02)            \\
\midrule
%\rowcolor[HTML]{FFFFFF} 
Adam                             & 71.16 (0.05)            & 70.60 (0.02)           \\
%\rowcolor[HTML]{FFFFFF} 
Adam-L                        & \textbf{76.18} (0.03)           & \textbf{76.07} (0.08)          \\
\midrule
%\rowcolor[HTML]{FFFFFF} 
LAMB+                            & 74.30 (0.01)               & 73.43 (0.03)              \\
%\rowcolor[HTML]{FFFFFF} 
LAMB-L                        & \textbf{76.48} (0.05)            & \textbf{75.93} (0.02)           \\
 \bottomrule
 \vspace{0.7cm}
\end{tabular}
}
      \captionof{table}{\textbf{ImageNet validation accuracy.} Comparison of LAWN variants (*-L) of different optimizers with their base variants on the ImageNet validation set. LAWN enables Adam to work on image classification tasks with very little drop in performance at large batch sizes. Current optimizers have a much steeper drop-off in performance as batch size increases. Standard error is mentioned in parentheses.}
      \label{tab:res-imagenet}
    \end{minipage}
  \end{minipage}

%We introduce a new method, Logit Attenuating Weight Normalization (LAWN), to handle the loss flattening and loss of adaptivity. The key idea is to not only constrain the weight norms of each layer for the network as training progresses but also use just the radial component of the gradient to update the parameters on each iteration. The LAWN update can be stacked on to any gradient-based optimizer where the training begins with unconstrained training using the base optimizer. After a few epochs, we the weight norms of the homogeneous layers are kept fixed at the respective weight norm values reached at that point. The rest of the training proceeds on the constrained norm surfaces employing projected gradients instead of the regular gradients. To demonstrate that LAWN makes the network more adaptive and helps escape points of poor generalization, we shrink the network weights of the aforementioned classification model at the point of peak generalization performance and continue training with the LAWN variant of Adam. As shown, the model escapes the poor minima and settles at a point of superior generalization with a significantly higher margin even though the training loss is not as low as it was earlier. This shows the power of LAWN in improving generalization performance and network adaptivity.

Consider that training has reached a point where logits are becoming large.
There are two ways of attenuating the logit values: (a) multiply the logits by a factor, $0 < \alpha < 1$ and use this factor for the rest of the training; (b) constrain the norms of layer weights for the rest of the training. 
%We introduce a new method, Logit Attenuating Weight Normalization (LAWN), to handle the issue of loss flattening and loss of adaptivity. The key idea is to constrain the weight norms of each layer of the network so that 
%the radial component of optimizer directions is removed and 
%the logits are controlled, thus increasing the adaptivity of the network needed to escape from the basins of sub-optimal minima. 
To provide a rough demonstration of the value of these methods, 
we return to the experiment of Figure~\ref{fig:lawn_motivation} and, 
after epoch $115$, we shrink the logits to one-fifth, and then continue training while keeping the weight norms of each layer fixed. This leads to superior generalization with no overfitting toward the end of training. Method (a) alone is not sufficient since further training will increase the logits again to reduce the training loss. Method (b) is more effective; also, in this case we could have worked with smaller logits by switching to constrained training very early.  
% also, it can work with smaller logits by switching to norm constraints early, say, after just 5 epochs in the example of Figure~\ref{fig:lawn_motivation}. \textcolor{orange}{[can we mention "work with smaller logits" in later section? this seems empirical experience]}

Our method, Logit Attenuating Weight Normalization (LAWN) can be used with any gradient-based base optimizer. LAWN begins with normal, unconstrained training in the initial phase and then fixes the weight norms of the layers for the rest of the training. The training on the constrained norm surfaces is done by employing projected gradients instead of regular gradients.

Although, adaptive methods like Adam have known to struggle with convolutional neural networks, the LAWN variant of Adam achieves very impressive performance across multiple network architectures and tasks. At large batch sizes, most optimizers get caught in sub-optimal regions due to lowered stochasticity which is worsened by increased logits. LAWN's attenuation of the logits helps avoid this worsening. Due to this, the LAWN variants of SGD, Adam and LAMB work significantly better than their base versions at large batch sizes. Table~\ref{tab:res-imagenet}, which compares the generalization performance of SGD, Adam and LAMB against their LAWN variants on the popular Imagenet dataset, powerfully showcases the two observations on LAWN mentioned above.

In the following sections, we motivate (\S\ref{sec:motivation}) and develop LAWN (\S\ref{sec:method}) as a new training method. We then proceed to show (\S\ref{sec:experiments}) that LAWN effectively improves adaptivity of networks and leads to much better generalization than known methods, e.g. $\ell_2$ regularization and weight decay using a wide variety of networks for image classification (CIFAR, ImageNet) and recommender systems (MovieLens, Pinterest).
%using a wide variety of networks for image classification (CIFAR, Imagenet) and recommender systems (MovieLens, Pinterest) that LAWN effectively improves adaptivity of the network and leads to much better generalization than known methods, e.g. $\ell_2$ regularization, weight decay, and label smoothing regularization. %

%While LAWN is particularly impressive in improving Adam, it greatly improves all optimizers when used with large batch sizes.

%In this paper we motivate and develop LAWN fully and demonstrate via detailed experiments on image classification datasets - {\em CIFAR}, {\em Imagenet} - and recommender datasets - {\em Movielens} and {\em Pinterest} that LAWN effectively improves adaptivity of the network and leads to much better generalization than L2 regularization and weight decay.

%\keerthi{I feel we should refer to Figure~\ref{fig:AlphaFactor} in the Intro and raise the interest of the readers. Plus, the contributions should come out well.}

%\end{document}
%\input{sections/related_work}

\def\to{\rightarrow}
\vspace{-0.25cm}
\section{The Need for LAWN}
\label{sec:motivation}
\vspace{-0.25cm}

In this section we motivate LAWN by describing the problem settings in which normal training struggles and in which LAWN could improve generalization performance. We define these problem settings in \S\ref{subsec:prsetting}, then describe what we mean by \textit{loss of adaptivity} in \S\ref{subsec:adloss}. We review the strengths and weaknesses of existing methods for avoiding loss of adaptivity in \S\ref{subsec:curradloss}. With this context, we elaborate on the LAWN method in \S\ref{sec:method}.

% {commented by Jun} We will start by defining these problem settings in subsection~\ref{subsec:prsetting}. As many deep net applications fall into these settings, LAWN can provide benefits in many situations. 
% \anika{In this section we motivate LAWN by describing and explaining the problem settings in which normal training struggles and in which LAWN could improve generalization performance. We will start by defining these problem settings in subsection~\ref{subsec:prsetting}. As many deep net applications fall into this settings, LAWN can provide benefits in most situations.} 

% {by Jun} A key reason why popular training methods such as SGD and Adam give sub-optimal performance is the loss of adaptivity of weights when training loss becomes small. This issue is discussed in detail in subsection~\ref{subsec:adloss}. In essence, LAWN is a training approach that can be used with any optimizer to overcome that issue; since different optimizers are suited for different application domains, e.g., SGD for Image classification, Adam for Recommender systems and NLP, the above property makes LAWN universally useful. There exist several other methods that are used to handle the issue of loss of adaptivity of weights. We review the strengths and weaknesses of some of them in subsection~\ref{subsec:curradloss}. With this context, we will describe and fully develop the LAWN method in section~\ref{sec:method}.

{\bf Notations.} For the rest of the paper we will use multi-class setting as the running example. The following notations will be used: $n$ is the number of weight variables;
$m$ is the number of training examples;
$i$ is the index used to denote the index of a training example;
$y_i$ is the target class of the $i$-th training example;
$k$ is the index used to denote a class; 
$nc$ is the number of classes;
$w$ is the weight vector of the deep net;
$p_k(w)$ is the class $k$ probability assigned by the deep net with weight vector $w$. For an optimizer, $\eta$ denotes the learning rate and $B$ is the batch size.

\vspace{-0.25cm}
\subsection{When does LAWN work?}
\label{subsec:prsetting}
% \anika{perhaps something more descriptive, ex: Problem Setting for Generalization Improvement via LAWN}\keerthi{Anika, That is too long a title for a subsection. But I have modified the first sentence below.}
%We now define notations used in the rest of the paper.

Let us describe the problem settings in which we expect generalization improvement using LAWN. 
%changed for to in%

{\bf Problem type.}
We are mainly interested in classification and ranking problems which form a score for the target. In binary classification this score is the logit score of the target class; in a multiclass problem, this score is the difference between the target class score and the maximum score over the remaining classes. For ease of presentation we will simply refer to such scores as {\em logit}. We consider a loss function that attains its least value (usually zero) asymptotically as logit goes to infinity. The logistic loss for binary classification and the cross entropy loss based on softmax for multi-class classification are important cases. Problems like regression that use a loss function such as the squared loss (which attains its minimum at a finite value) may not have much benefit using LAWN. In a multi-task setting, when the total loss is an additive mix of several individual task losses, LAWN can be useful even if just one of them is suited for LAWN.

{\bf Homogeneity.} A network layer is said to be {\it homogeneous} if %multiplying all weights of the layer 
multiplying either all weights or all incoming signals of the layer by a constant $\alpha$ leads to the output of the layer being multiplied by $\alpha$.
%\footnote{We can be more general and consider higher degree by replacing the output factor to be $\alpha^p$ for some positive integer $p$. For simplicity we restrict ourselves to degree $1$.} 
The presence of bias weights in a layer causes a violation of homogeneity since the input multiplied by them is always $1$, i.e., $W(\alpha x)+b \not= \alpha (Wx+b)$. Since the number of bias weights in any layer is small, we will ignore this as a minor issue and take a layer (which is otherwise homogeneous) to be homogeneous for all theoretical reasoning; also, see Lyu and Li~\citep{lyu2020gradient} for a discussion of this matter. 
A layer of {\em linear} or {\em relu} units is homogeneous. When classification and ranking problems are used, the final layer of the network is usually linear (it generates the class scores) and hence homogeneous. Homogeneity in the end layer(s) is the reason for logit to become large and hence it plays a key role in the utility of LAWN.
A maximal contiguous set of 
%\anika{homogenous} 
homogeneous layers of a network will be referred to as a {\em homogeneous subnet}. 
%In a homogeneous subnet with $j$ layers, when all weights of the subnet are multiplied by $\alpha$, does the output get multiplied by $\alpha^j$
%where $j$ is the number of layers in the subnet%
%? Not necessarily; the caveat being the presence of bias weights in the layers since the input weight received by them is always $1$. \textcolor{red}{[Jun] Confused, does homogeneous layer have bias or not?} %Since the number of bias weights is small,we can ignore this as a minor issue. For theoretical reasoning\textcolor{orange}{,} we will assume the $\alpha^j$ multiplicative property to hold. 

\def\fhsn{{\em fhsn}}
The final layer that forms the class scores and all preceding contiguous homogenous layers is called the final homogeneous subnet ({\em fhsn}). In the case of a batch normalization (BN) layer, we can break the layer into two sublayers: the first sublayer does normalization (non-homogeneous) over a batch of examples, and the following second sublayer applies scaling (homogeneous) and  shift (which is akin to bias weights). Thus, when BN is encountered, the second sublayer is treated as homogeneous and, if it is immediately followed by a homogeneous activation function, then it gets included as a part of the homogeneous subnet that follows it. Similar ideas hold for other normalizations such as layer normalization. We will refer to the weights in \fhsn~as $w_{fhsn}$. Figure~\ref{fig:fhsnexample} gives an example to illustrate these definitions.

%The final homogeneous subnet ({\em fhsn}) of a model includes the final layer that forms the class scores and all preceding contiguous homogenous layers. 

% \begin{figure}[h]
%     \begin{center}
%         \includegraphics[scale=0.4]{images/DefineHomogenousSubnet.png}
%     \end{center}
%     \caption{An example that depicts the definitions of a {\em homogeneous subnet} and \fhsn.}
%     \label{fig:fhsnexample}
% \end{figure}

\begin{wrapfigure}{R}{4cm}
\centering
\vspace{-1cm}
\includegraphics[scale=0.3]{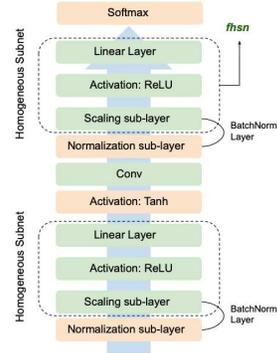}
\caption{Homogeneous layers are shown in green, with each homogeneous subnet marked in dotted lines. \fhsn~is the final homogeneous subnet.}
\label{fig:fhsnexample}
\vspace{-1cm}
\end{wrapfigure}

% \begin{wrapfigure}{R}{6.75cm}
% \centering
%     \includegraphics[width=0.36\columnwidth]{images/constrainedstepofsgd.jpg}
%     \vspace{-0.2cm}
%     \caption{A constrained training step of SGD-LAWN.}
%     \label{fig:sgdlawn}
% \vskip -0.1in
% \end{wrapfigure}

% \begin{figure}
% \vspace{-0.1cm}
% \begin{center}
% \centerline{\includegraphics[width=0.4\columnwidth,trim={1cm 5cm 12cm 1cm}, clip]{images/network_arch.pdf}}
% \caption{Homogeneous layers are shown in green, with each homogeneous subnet marked in dotted lines. The final homogeneous subnet is marked as \fhsn}
% \label{fig:fhsnexample}
% \end{center}
% \vspace{-1cm}
% \end{figure}

When \fhsn~contains all the layers of the network, we call it a {\em fully homogeneous} network. Fully homogeneous networks are popularly used in recommender systems~\citep{he2017neural, zhang2019deeprecsyscsur}.

\textbf{Network Complexity.} This refers to the complexity of the network in relation to the number of training examples. Most deep networks used in applications are {\em over-parametrized}. Roughly, we will take {\em over-parametrized} to mean that the network is so powerful that training easily locates a $w$ that classifies all examples correctly, i.e., the target class has the highest score among all classes. Given the presence of \fhsn~which can lead to large logit scores, this would also mean that, by making $w_{fhsn}$ go to $\infty$ we can make the training loss go to zero.

For LAWN to work usefully, we do not require over-parametrization per se, but a network which is very flexible and training easily pushes the loss of most training examples very small. A small set of examples can remain misclassified; for example, this often happens when data augmentation is used. 
Below, in \S\ref{subsec:adloss} we will explain how this situation of most examples attaining very small loss values creates an issue.

% Here it is worth mentioning about the phenomenon of double descent in which, as network complexity is increased, generalization error reaches a dip, then increases, reaches a peak, and then turns down to reach the best generalization error when the network complexity is very large. \keerthi{Aman questioned this. Let's look at some DD papers and discuss this.} This final and important zone is ideal for LAWN to yield further improvement in generalization error over a normal unregularized training method. \keerthi{Aman, If there is time, it would be useful to include one experiment on this.}

\textbf{Generalization Metrics.} Our focus is on improving metrics that are based on score ordering as opposed to the actual values of scores. More precisely, we are interested in test set metrics such as classification error, AUC, NDCG etc. as opposed to logistic loss, cross entropy loss, probability calibration error etc. Deep networks are known to be poor with respect to the latter metrics~\citep{GuoPSW17} but which can be improved in the post-training stage; the adaptation of LAWN to improving such metrics will be taken up in a follow-up work. 
% \textcolor{red}{[mizhou]which are popular metrics works well for deep networks}

\subsection{Issue of loss of adaptivity}
\label{subsec:adloss}

Consider the problem setting defined in \S\ref{subsec:prsetting} and the minimization of the normally used unregularized training loss,
$L(w)  =  -\frac{1}{m}\sum_{i=1}^m \log p_{y_i}(w)$. 
With the network being sufficiently powerful, training causes the losses of a large fraction of the training examples to become very small. During training, this happens due to $w_{fhsn}$ becoming large, the logit becoming large, and $p_{y_i}\to 1$ for those examples. Due to these, each of the following become very small: the gradient of most example-wise loss terms; $\Sigma$, the covariance of the noise associated with minibatch gradient; and, $H$, the Hessian of the training loss. We will refer to this collective happening as {\em loss flattening}.

It has been established via theoretical and empirical arguments that the powerful generalization ability of a deep net comes from its ability to escape regions of attraction of ``sharper" minima\footnote{Originating from the work of Keskar et al~\citep{keskar2016large}, it has been observed in the literature that minima at which the classification function has a sharper behavior has poorer generalization.} with sub-optimal generalization performance and go to better solutions. Appendix~\ref{sec:escape} gives an idea of the escape mechanism for the SGD method using an analysis given by Wu et al~\citep{Wu_NEURIPS2018}. It is worth recalling from there, the following rough guiding condition for SGD to escape from a poor solution: $\lambda_{\max} \{ (I-\eta H)^2 + \frac{\eta^2(m-B)}{B(m-1)} \Sigma \} > 1$, 
%\label{eq:Wu}
where $\lambda_{\max}(A)$ denotes the largest eigenvalue of $A$. If training is at a sub-optimal generalization point and loss flattening occurs (which means that $H$ and $\Sigma$ become small), then the escape condition 
%in (\ref{eq:Wu}) 
is difficult to satisfy and hence it becomes difficult for the network to escape from this solution and then train further to go to a better solution. A carefully increased learning rate schedule to cause the escape followed by the use of normal learning rates to go to a better solution can make this happen, but no such automatic sophisticated learning rate adjustment mechanism has been devised yet. We will refer to this inability of the network to escape out of a sub-optimal solution due to loss flattening as {\em loss of adaptivity} (also see \S7 in ~\cite{szegedy2016rethinking}).

\def\elsr{\epsilon_{LSR}}
\def\ef{\epsilon_{Flooding}}

\subsection{Current methods for dealing with loss of adaptivity}
\label{subsec:curradloss}
Several methods have been suggested in the literature to handle the issue of loss of adaptivity. We briefly describe three key ones: label smoothing regularization (LSR)~\citep{szegedy2016rethinking}, flooding~\citep{ishida2020we}, and $\ell_2$ regularization/weight decay~\citep{loshchilov2019decoupled}. 
LSR modifies $L(w)$ via making the target class less confident by fixing its probability as $(1-\elsr)$ and reassigning $\elsr$ to the remaining classes. This makes the loss attain its minimum at finite logit values. Flooding modifies the loss as $L_{Flooding}(w) = |L \; - \; \ef|$ so that the training process is forced to move around the hypersurface defined by $L \; - \; \ef = 0$. $\ell_2$ regularization is a traditional method that modifies the loss as $L_{\ell_2} = L(w) + \frac{\lambda}{2} \|w\|^2$. Weight decay, as used in the recent deep net literature decouples the term $\lambda w$ from the gradient and instead includes the additive term, $-\lambda w$ at the weight update step.
Later, in \S\ref{sec:method} we will return to discuss these methods in relation to LAWN.

\vspace{-0.25cm}
\section{The LAWN Method}
\label{sec:method}
\vspace{-0.25cm}

\def\to{\rightarrow}
\def\wbar{\bar{w}}
\def\fhsn{{\em fhsn}}

% \begin{wraptable}{L}{7cm}
% \centering
% \resizebox{0.48\textwidth}{!}{%
% \begin{tabular}{lllll}
% \toprule
% Batch Size & LSR   & Flooding & WD & LAWN \\ \midrule
% 10k & 66.57 & 65.51    &70.6    &70.9 \\
% 100k & 66.12 & 64.29   &67.2    &70.3 \\ 
% \bottomrule
% \end{tabular}
% }
% \caption{LAWN vs other methods for controlling loss of adaptivity. For two different batch sizes, LAWN comprehensively outperforms other methods on a recommendation systems task. \textcolor{red}{Mention that the poor methods will not be used again.}}
% \vspace{-3mm}
% \label{tab:lawn-vs-other}
% \end{wraptable}

A smooth classification loss such as the one based on logistic or cross entropy functions is necessary for gradient-based training. Consider a network with weights $\wbar$ and corresponding \fhsn~weights $\wbar_{fhsn}$. When we extend $\wbar_{fhsn}$ along a radial direction, i.e., $w_{fhsn} = \alpha \wbar_{fhsn}$, $\alpha\in R_+$, while keeping the weights of the network outside \fhsn~unchanged, the classification error on a set of examples (e.g., the training error computed on a training set or the generalization error computed on a test set) remains unaffected whereas classification loss varies a lot with $\alpha$. In particular, if $\wbar$ classifies a set of examples strictly correctly, then, as $\alpha$ goes from $0$ to $\infty$, the average loss on these examples goes from $\log (nc)$ to zero asymptotically. During normal training, with weight directions having a good radial component, training loss does become very small. Earlier, in \S\ref{sec:motivation} we saw how, when training loss becomes very small, it causes loss flattening, which leads to a loss of adaptivity of the network. Therefore, it makes sense to suitably contain the radial movement of $w_{fhsn}$ by constraining $\|w_{fhsn}\|$. The essential spirit of LAWN is along this idea.\footnote{Weight norm bounding in neural nets is theoretically well-founded for improving generalization~\cite{NeyshaburTS15, BartlettFT17}; in LAWN, the bounding is done in a specific way to avoid loss of adaptivity.} 

\fhsn~is a homogeneous subnet which can have more than one layer. A homogeneous subnet has many redundancies~\citep{dinh2017sharp}. For example, we can take any one layer, multiply all weights in the layer by a constant and divide all the weights of another layer by the same constant and hence keep the input-output function of the subnet unaltered. Given this, one can choose to constrain the weights in \fhsn~in one of the following two ways: (a) coarsely at the full \fhsn~level by constraining  $\|w_{fhsn}\|$; (b) finely at the layer level by constraining each $\| w_{fhsn}^{\ell}\|$ where $w_{fhsn}^{\ell}$ denotes the weights of a layer $\ell$ in \fhsn. In this paper we work with constraints at the layer level since it removes more redundancies.

%\fhsn~is a homogeneous subnet which can have more than one layer. A homogeneous subnet has many redundancies~\citep{dinh2017sharp}. For example, we can take any unit (neuron) of one layer, multiply all the incoming weights to its activation function by a constant and divide all its outgoing weights (the weights that take this unit's output as their input) by the same constant and hence keep the input-output function of the subnet unaltered. Given this, one can choose to constrain the weights in \fhsn~in one of the following three ways, going from coarse to fine to very fine: (a) at the full \fhsn~level by constraining  $\|w_{fhsn}\|$; (b) at the layer level by constraining each $\| w_{fhsn}^{\ell}\|$ where $w_{fhsn}^{\ell}$ denotes the weights of a layer $\ell$ in \fhsn; (c) at the unit level by constraining each $\| w_{fhsn}^i\|$ where $w_{fhsn}^i$ denotes the weights of a unit $i$ in \fhsn. In this paper we take the middle path and operate at the layer level.

Consider constrained training via layer-wise weight normalization: $\| w_{fhsn}^{\ell}\| = c^{\ell} \; \forall \ell$ where the $c^{\ell}$ are kept constant. It is useful to understand how the loss contours behave as the $c^{\ell}$ go from small to big values. When the $c^{\ell}$ are large, we know that loss flattening will happen. When the $c^{\ell}$ are small, the distinction between the loss values of well classified examples and poorly classified examples diminishes and so, optimizers will find it harder to traverse the contours and go to the right place of best generalization. Given this, we take a simple and natural approach to LAWN training. We initialize the network with weights having small magnitude using a standard weight initialization method and start a given optimizer in its free (unconstrained) form. At a suitable point in that training process, with weights at some $\wbar$, we switch to constrained training defined by setting
\begin{equation}
    \|w_{fhsn}^\ell \| = c^{\ell} \mbox{ where } c^{\ell} = \| \wbar_{fhsn}^{\ell} \| \;\; \forall \ell \label{eq:constrdef}
\end{equation} 
The $c^{\ell}$ are fixed for the rest of the LAWN training. Constrained training corresponds to solving the optimization problem,
\begin{equation}
    \min L(w) \mbox{   s.t.   (\ref{eq:constrdef}) }
    \label{eq:constropt}
\end{equation}
using a modified version of any given gradient-based optimizer. 

Though the homogeneous layers outside \fhsn~do not contribute to the growing of the scores at the end of the network that leads to loss flattening, for better performance we have found that it is useful to constrain them too. In the case of BN, we can even constrain the weight layer that just precedes it. Thus, in LAWN, the $\ell$ in (\ref{eq:constrdef}) can be taken to include all such weight groups in the network;
%\footnote{Note that any normalization in an extended group, being homogeneous, is either adjusted or absorbed in one of the following layers and so the capability of the network is unaffected.}
we do this in our implementation and use it for all experiments reported in this paper.
%and we refer to this as {\em ehl}. 
%Table~\ref{tab:cifar-fhsn-ahl} compares \fhsn~ and {\em ehl} on the CIFAR datasets. 
%Since {\em ehl} is better, we use it in our implementation and for all experiments reported in this paper.

%\begin{table}[!ht]
%\centering
%\resizebox{0.75\textwidth}{!}{%
%\begin{tabular}{c|c|c|c|c}
%  \multicolumn{1}{c}{Adam-LAWN} &
%  \multicolumn{2}{c}{CIFAR10} &
%  \multicolumn{2}{c}{CIFAR100} \\
%\hline
%batch size & fhsn & ehl & fhsn & ehl \\
% \hline
% \multirow{3}{*}{SGD-LAWN} & 256 & 94.04 & 94.34 & 70.24 & 71.11 \\ \cline{2-6}
%                      & 4k & 92.8 & 92.79 & 68.3 & 69.15 \\ \cline{2-6}
%                      & 10k & 91.69 & 92.06 & 66.05 & 66.53 \\
%\hline
%\multirow{3}{*} 256 & 93.9 & 94.15 & 71.1 & 73.47 \\ \cline{2-6}
%                     4k & 93.31 & 94.23 & 70.43 & 73.88 \\ %\cline{2-6}
%                     10k & 92.97 & 94.03 & 68.96 & 73.31 \\
%\hline
% \multirow{3}{*}{Lamb-LAWN} & 256 & 94.02 & 94.13 & 71.32 & 72.07 \\ \cline{2-6}
%                      & 4k & 93.74 & 93.83 & 71.43 & 70.33 \\ \cline{2-6}
%                      & 10k & 93.43 & 93.5 & 70.21 & 69.83 \\ 
% \hline
%\end{tabular}
%}
%\caption{Comparing LAWN applied to the end homogeneous subnet (fhsn) and to the extended homogeneous layers (ehl) for CIFAR10 and CIFAR100. Clearly, ehl is better.}
%\label{tab:cifar-fhsn-ahl}
%\end{table}

%\textcolor{red}{[Jun] The next sentence is long-winded. Try this: 
The switching from free training to constrained training can be done by one of the following methods.
%} 
%The point in free training at which switching is done from free to constrained training can be done in one of two ways. 
(1) We can switch after running free training for a certain number of epochs, and tune this number as a hyperparameter using a coarse grid. 
%Just a few epochs of free training is usually sufficient for LAWN to switch to constrained training and generalize well. 
(2) We can track the training accuracy (done simply and efficiently by tracking accuracy on the minibatches used) and switch to constrained training when it starts plateauing at a decent high value and still far from loss flattening. All experiments reported in this paper are done using the first method. It would be useful to try the second method in the future.

%\keerthi{Let's add 3 references from reviewer 1 of the FixNorm paper - https://openreview.net/forum?id=pD7kGiAQkNY. We can talk about weight norm being good, and how we use it for a specific purpose, which is preventing loss flattening. }
Hoffer et al~\cite{hoffer2018norm} gives a bounded weight normalization method which also constrains the weight norm. However, 
%it normalizes weights at the unit (neuron) level, it uses the idea only for a fully homogeneous setting, 
it does not combine free and constrained trainings, and it does not set up and demonstrate the use of constrained training as a powerful method for use with a general optimizer for improving the performance by overcoming loss flattening and loss of adaptivity. Also, the method, as stated in the paper, does not work well. See Table~\ref{tab:lawn-vs-other}.

%\subsection{LAWN Implementation}
%\label{subsec:lawnimp}
%\keerthi{We need to carefully talk about why the learning rate schedule is 3-phase. LAWN uses two different kinds of training, so careful schedule is important (2 warm ups needed). }
%\begin{wrapfigure}{L}{0.7\textwidth}
\begin{wrapfigure}{L}{0.66\textwidth}
\vspace{-0.8cm}
\begin{minipage}{0.66\textwidth}
    \begin{algorithm}[H]
    \caption{LAWN Method}
    \label{algo:lawn}
    %\DontPrintSemicolon
    \begin{algorithmic}[1]
    
    \State Fix $E_{total}$ and {\em BatchSize} to suitable values.
    \State Choose a set of values for $\eta_{peak}$, $E_{warmup}$ and $E_{free}$.
    
    \For {each set of hyperparameter choices}
        \State Initialize the network weights.
        \State Start the free (unconstrained) training with  linear warmup.
        \State Stop after $E_{free}$ epochs and switch to constrained training.
        \State Do constrained training with linear warmup and decay.
    \EndFor
    
    \end{algorithmic}
    \end{algorithm}
\end{minipage}
\vspace{-0.5cm}
\end{wrapfigure}
%\end{wrapfigure}

        % \Comment{\textcolor{red}{\small Free training will be done as if it will go for $E_{total}$ epochs. But it will be cut short after $E_{free}$ epochs.}} 
\textbf{LAWN Implementation.} We employ a simple approach to the implementation of constrained training. Let $x$ denote a weight subvector of any group (for eg., the weights and biases of a fully connected layer constitute one group) and some generic optimizer suggests a direction $d$ for $x$ to move. To remove the radial component we modify $d\to d_p$ by projecting $d$ to the hyperplane with $x$ as its normal. Then we do the update $x_{new} = x + \eta d_p$ where $\eta$ is the learning rate. Since this step increases the norm, we complete the step with a normalization: $x_{new} \leftarrow \beta x_{new}$ where $\beta = \|x\|/\|x_{new}\|$. If an adaptive optimizer such as Adam is used, all direction-related subvectors used are changed to the corresponding projected quantities; in particular, if $g^x$ is the gradient subvector corresponding to $x$ then the projection of $g^x$ to the hyperplane with $x$ as its normal is used in all updates involving the gradient. Appendix~\ref{app:lawnalgos} gives the full details of SGD-LAWN, Adam-LAWN and LAMB-LAWN.

% \textcolor{red}{[Jun] are we going to list all three in the Appendix?}) 
% Figure~\ref{fig:sgdlawn} illustrates the steps for SGD-LAWN.
%\keerthi{Anika to remove Wl and change it to x to be consistent with the text.}

% \begin{wrapfigure}{R}{4.5cm}
% \centering
% \vspace{-0.5cm}

%     \includegraphics[width=0.36\columnwidth]{images/constrainedstepofsgd.jpg}
%     \vspace{-0.2cm}
%     \caption{\textit{A constrained training step of SGD-LAWN.}}
%     \label{fig:sgdlawn}
% \vskip -0.1in
% \end{wrapfigure}

% \begin{figure}[h]
%     \centering
%     \includegraphics[width=0.5\columnwidth]{images/constrainedstepofsgd.jpg}
%     \vspace{-0.2cm}
%     \caption{A constrained training step of SGD-LAWN.}
%     \label{fig:sgdlawn}
% \vskip -0.1in
% \end{figure}

An alternative implementation to deal with the norm constraint, $\|x\|=c$ is to define an unconstrained vector $v$ and set 
$x=c\;v/||v||$ and simply apply any optimizer to $v$. This is the implementation suggested in Salimans and Kingma~\citep{salimans2016weight}; see equation (2) there. (Note, however, that Salimans and Kingma~\citep{salimans2016weight} keeps the radial component by including the scale parameter, $g$.) One downside with this method, which makes normalization a part of the computational graph, is that
automatic differentiation through the normalizing transformation increases
the computational cost~\cite{Huang2021}. It is worth considering this idea in future implementations of LAWN.

Let us now write the LAWN method as a complete algorithm. The free and constrained trainings of LAWN can be done with any given optimizer. LAWN does a total of $E_{total}$ epochs; $E_{total}$ is fixed for a given dataset. After $E_{free}$ epochs of free training it switches to do constrained training for $(E_{total}-E_{free})$ epochs. Learning rate schedules are important for deep networks to attain good performance~\citep{loshchilov2019decoupled, liu2020variance, loshchilov2016sgdr}. 
In free training we linearly ramp the learning rate from $0$ to $\eta_{peak}$ in $E_{free}$ epochs. In constrained training, we ramp the learning rate from $0$ to $\eta_{peak}$ in $E_{warmup}$ epochs and then linearly ramp down the learning rate from $\eta_{peak}$ to $0$ in the remaining epochs. Figure \ref{lrsched_lawn} in Appendix~\ref{app:experiments} shows this in detail. 
The LAWN algorithm is described in Algorithm~\ref{algo:lawn}. 

%For LAWN, we employ warmup and decay in the free\footnote{Usually $E_{free}$ is so small that the decay part of free training is rarely encountered.} and constrained training phases using linear learning rate schedules: In $E_{warmup}$ epochs $\eta$ is changed from zero to $\eta_{peak}$; Then $\eta$ is decreased from $\eta_{peak}$ to zero when a total of $E{total}$ epochs is reached. 
% We tune $E_{free}$ and $E_{warmup}$ over a coarse grid of one to a few epochs, and $\eta_{peak}$ over a logarithmic grid of ten values. 

% When batchsize is changed, it would be useful to say if there are ways of setting  $\eta_{peak}$, $E_{warmup}$ and $E_{free}$ for the new batchsize based on the optimal values chosen for the previous batchsize. \keerthi{Aman, Jun, we need to discuss how to write this. If there is no clear conclusion, let's leave it out.}

%\subsection{Fully homogeneous nets and implicit bias}
%\label{subsec:impbias}

%\keerthi{This section needs to be re-written, the content around eq 9 to 10.}

{\bf Fully homogeneous nets and implicit bias.}
Suppose the network that we use is fully homogeneous and we have an over-parametrized training problem. For such classification problems, it has been shown that gradient descent methods~\citep{lyu2020gradient, Poggio2019, Poggio2020, Wang2020, soudry2018implicit} have good implicit (inductive) bias properties.
%\footnote{Implicit bias is a (good) property that is unintended in the optimization formulation, but which happens because of the optimization method used to solve that formulation.} 
Specifically, it has been shown for fully homogeneous nets that gradient-based methods asymptotically attain weight directions that maximize the normalized margin (``max margin"):
\begin{equation}
    \max_w \mbox{ margin}(w) / \Pi_{\ell} \|w^{\ell}\| \label{eq:normmarginmax}
\end{equation}
where $\mbox{margin}(w)$ is the margin given by the network with weights $w$ on the training set. Clearly, for any $\delta^\ell \in R_+$, doing $w^\ell \leftarrow \delta^\ell w^\ell$ individually for each $\ell$ will not alter the value of the normalized margin.
So, it is not surprising that the max margin directions can also be obtained by solving~\citep{Banburski2019}
\begin{equation}
    \max_w \mbox{ margin}(w) \mbox{. s.t.  } \|w_{fhsn}^\ell \| = c^{\ell} \;\; \forall \ell \label{eq:constrmaxmargin}
\end{equation}
See Appendix~\ref{sec:marginproof} for a proof of this result. 
% \textcolor{red}{[Jun] discussion: can $c^{\ell}$ be one?}

% {\bf Experiment.} Experiment on Movielens 100k studying the correlation between normalized margin and generalization error. \keerthi{We can leave this out.}

Though there have been attempts at optimizing margin maximization directly~\citep{bansal2018minnorm}, it is not worth attempting because: (a) for problems that are mildly over-parametrized, it can be very sensitive to noise and lead to poor generalization and (b) the severe non-convexity and added non-smoothness of the objective make it a very hard problem to solve.

% \footnote{Aman, I think this can explain the final drop in generalization in Movielens. Anika can do a simple 2d experiment to show this. We can discuss if it is worth including in an appendix. This also has connections to double descent.}

A better approach is to blend naturally with normal training via approximating $\mbox{ margin}(w)$ by a smooth function such as $h(L_{Normal}(w))$ where $h(x) = -\log (exp(x)-1)$. Thus, (\ref{eq:normmarginmax}) can be effectively approximated by
\begin{equation}
    \max_w h(L(w)) / \Pi_{\ell} \|w^{\ell}\| \label{eq:appnormmarginmax}
\end{equation}
Lyu and Li~\citep{lyu2020gradient} makes effective use of (\ref{eq:appnormmarginmax}) in its theory of implicit bias. Since $h(x)$ decreases monotonously for positive $x$, maximizing $\mbox{ margin}(w)$ is equivalent to minimizing $L$ and this leads us directly to (\ref{eq:constropt}) as an approximation of (\ref{eq:constrmaxmargin}). This is a great property of LAWN - that, its constrained optimization problem, (\ref{eq:constropt}), is closely connected to the implicit bias defined by (\ref{eq:normmarginmax}). As the $\{c^{\ell}\}$ become large, the solutions of (\ref{eq:constropt}) converge in direction to the solutions of (\ref{eq:normmarginmax}). Thus we can view LAWN's constrained optimization as a way of capturing implicit bias while also avoiding loss flattening. Poggio et al~\citep{Poggio2019, Poggio2020} discuss these properties for fully homogeneous nets, but they do not point out the issue of loss flattening and loss of adaptivity and so they do not develop an effective overall training method like LAWN to overcome that issue.

LSR and Flooding do not enjoy the implicit bias property because their optima are finitely located. By themselves, they have to mainly rely on the noise associated with minibatch stochasticity for reaching points of good generalization.\footnote{It is worth noting that Szegedy et al~\citep{szegedy2016rethinking}, which introduced the idea of LSR, also uses $\ell_2$ regularization for training.} On the other hand, $\ell_2$ regularization, like LAWN, also approximates implicit bias - the former uses the weight norm in the objective function and the latter uses it in the constraints. See Appendix~\ref{sec:marginproof} for details. Having said that, (\ref{eq:constropt}) used by LAWN could be better because of the following reasons. (a) It does not interfere with the formation of the main classification boundary in the early phase of free training. (b) The $c^{\ell}$ values are set naturally and differently for different layers of the network and so they do not require tuning. On the other hand, the $\ell_2$ regularization constant, $\rho$ is a single value for all the layers and it needs to be chosen to avoid the extremes of loss flattening (which happens if $\rho$ is set too low) and too much interference with loss minimization (which happens if $\rho$ is too big). In general, $\rho$ has to be carefully tuned, say using a logarithmically spaced set of values, to get the best performance ~\citep{hanson1988comparing, loshchilov2019decoupled}.
%\keerthi{The following will be rewritten after Aman completes the experiments.}
%The following two experiments are worth adding here.
%\begin{enumerate}
    %\item Experiment on Movielens 100k studying normalized margin and its correlation with generalization error. \keerthi{Aman to run this}
    %\item On the same dataset, show how normalized margin and gen perf change with the $\alpha$ factor - to show that the approximation of implicit bias via logistic/cross entropy is good. \keerthi{Keerthi to define experiment.}
%\end{enumerate}

\begin{wraptable}{L}{7cm}
\centering
\resizebox{0.5\textwidth}{!}{%
\begin{tabular}{llllll}
\toprule
\textbf{Batch Size} & \textbf{LSR}   & \textbf{Flooding} & \textbf{WD} & \textbf{Hoffer} & \textbf{LAWN} \\ \midrule
10k & 68.66 & 68.44    &70.12 &44.54   & {\bf 70.41} \\
100k & 67.34 & 67.60   &69.28 &45.52   & {\bf 70.77} \\ 
\bottomrule
\end{tabular}
}
\caption{Test HR@10 (Movielens-1M) of LAWN vs. other methods for controlling loss of adaptivity. For two different batch sizes, LAWN comprehensively outperforms other methods, including Weight Decay (WD).
%on a recommendation systems task. 
Complete details are in Appendix~\ref{app:experiments}.}
\vspace{-4mm}
\label{tab:lawn-vs-other}
\end{wraptable}

Though implicit bias has been proved only for fully homogeneous networks, one expects to see some form of max margin theory of implicit bias for non-homogeneous nets having \fhsn~ to be demonstrated in the future. \fhsn~is expected to play a key role in this theory. Irrespective of such a theory, the use of constrained optimization in LAWN is well-motivated for such general nets since controlling radial movements of \fhsn~ avoids loss flattening and loss of adaptivity.

%\subsection{LAWN and large batchsizes}
%\label{subsec:largebatch}
% \keerthi{move the one-off experiment here.}
%\keerthi{Probably needs revision.}
{\bf Quick comparison with other methods.}
We used a recommendation systems task (MovieLens-1M classification) to compare LAWN to other methods for controlling loss of adaptivity across two different batch sizes. Table~\ref{tab:lawn-vs-other} shows the results. LAWN outperforms all other methods. The second best method is weight decay and hence it is used as the baseline for all remaining experiments. The weight normalization technique suggested by Hoffer et al~\citep{hoffer2018norm} requires a modification along the lines of LAWN in order for it do well.

{\bf LAWN and large batch sizes.}
For a fixed epoch budget, larger batch sizes require a smaller number of steps; combined with distributed computation this helps speed up training. However, since stochasticity of updates reduces with large batch size, the mechanism of escape from sub-optimal solutions gets affected. Thus, one usually sees a reduction in generalization performance as batch size is increased~\citep{shallue2018measuring}. Loss flattening makes this issue worse by affecting adaptivity. LAWN, by helping overcome this issue, leads to a more graceful degradation of performance as a function of batch size. The degradation becomes far less (even zero) when the the total number of steps is allowed to decently increase with batch size.

%This paragraph needs to be expanded and re-written. Cite the JMLR paper (section 5) shared by Keerthi. Also point to evidence from our experiments on

%\keerthi{Should we tak about expanding step budget for large batch sizes?}

%Large batch sizes are useful in speeding up training if compute is not a constraint. However, since stochasticity of updates reduces with large batchsize, the mechanism of escape from sub-optimal solutions gets affected. Thus, one usually sees a reduction in generalization performance as batchsize is increased beyond a certain value. But, as we will show in the experiments section, LAWN maintains retains strong generalization performance for ImageNet classification and recommender systems even for large batch sizes.

%\keerthi{LAWN helps avoid loss flattening *does not mean* that it overcomes the inability of large batch sizes to generalize well. This is because, even with LAWN, the noise needed to escape poor solutions can be low as batch is increased. We should not claim or even give the impression that LAWN will make things great for any value of large batch. Having said that, we should also stress the Movielens case, where good generalization does happen with LAWN even for very large batch sizes. TODO address the comment where we discuss large batch sizes. Graceful degradation.}

%\subsection{An extended LAWN method}
%\label{subsec:alphatrick}

%\keerthi{This subsection has been rewritten.}

{\bf An extended LAWN method.} 
Suppose we keep track of a historically smoothed average logit score over an epoch, and at some point of the training find that it exceeds a threshold that is set to a value around the point where loss flattens, say 3 for logistic loss (see Appendix~\ref{sec:lossflattening} for a plot of the logistic loss function). 
We can multiply the class scores of the network by a factor, $0 < \alpha < 1$ and do constrained LAWN training while also multiplying the class scores by the same fixed factor $\alpha$ for the rest of the training. Earlier, in \S\ref{sec:intro}, Figure~\ref{fig:lawn_motivation} we demonstrated the value of this simple method for a recommender network on the Movielens datset. This can lead to an automatic way for deciding when to switch from free to constrained training. We have not used this idea in the experiments of \S\ref{sec:experiments}. 

\vspace{-0.5cm}
\section{Experiments}
\label{sec:experiments}
\vspace{-0.25cm}

To evaluate LAWN, we compared regular and LAWN-based variants of some optimizers - SGD with momentum~\citep{qian1999momentum}, Adam~\citep{kingma2017adam} and LAMB~\citep{you2020large}. Traditionally, SGD with momentum has demonstrated strong performance for computer vision tasks ~\citep{ren2015faster, goyal2018accurate}, whereas adaptive methods like Adam perform well on other domains (eg. recommender systems, text classification). To demonstrate the efficacy of LAWN across a wide variety of tasks, we conducted experiments on the CIFAR~\citep{krizhevsky2009cifar} and ImageNet~\citep{deng2009imagenet, krizhevsky2012imagenet} datasets for image classification, and the MovieLens~\citep{harper2015movielens} and Pinterest~\citep{geng2015learning} datasets for item recommendation. All model training code was implemented using the PyTorch library~\citep{NEURIPS2019_9015} and experiments were conducted on machines with NVIDIA V100 GPUs. For each experiment, we report the average test metric over 3 runs. We did not tune the Adam hyperparameters $\epsilon$, $\beta_1$ and $\beta_2$, which could have led to further improvements for Adam-LAWN.

All LAWN optimizers use a 3-phase learning rate schedule as described in \S\ref{sec:method}. Base variants use a 2-phase learning rate schedule that incorporates both warmup and decay. For base variants, we tuned $E_{warmup}$, peak learning rate, and weight decay. For LAWN variants, we tuned $E_{free}$, $E_{warmup}$ and peak learning rate. Appendix \ref{app:experiments} contains details about hyperparameters and their tuning.

\subsection{Image Classification for CIFAR-10 and CIFAR-100}

% \keerthi{Need to carefully explain learning rate scheduling here. Explain different learning rates for different phases. Aman - I have already done that in detail in the appendix}
For both CIFAR-10 and CIFAR-100~\citep{krizhevsky2009cifar}, we used the VGG-19 CNN  network~\citep{simonyan2014very} with 1 fully connected final layer. Our ImageNet experiments use a ResNet-based~\citep{he2015deep} architecture. All experiments were run with a 300 epoch budget.
%\textbf{Overall Results.} From Table \ref{tab:img_result}, it is evident that 
As seen in Table~\ref{tab:img_result}, LAWN variants either match or outperform the base variants across batch sizes. Adam-LAWN is particularly impressive. This is in stark contrast to earlier held beliefs that adaptive optimizers cannot match SGD's generalization performance for image classification tasks~\citep{wilson2017marginal}.

\begin{figure*}[htb!]
\centering
\subfigure[CIFAR-10]{%
    \pgfplotsset{every axis/.append style={
        font=\fontsize{16}{16}\selectfont,
        line width=2pt,
        mark size=2.5pt,
        tick style={line width=0.8pt}
    }}
    \begin{tikzpicture}[scale=0.35]
    	\begin{semilogxaxis}[
		    xlabel=Batch size,ylabel=Test Accuracy,
		    ymin=89, ymax=95,
		    ymajorgrids,
		    xmajorgrids,
		    legend style={at={(0.03,0.03)},
		    anchor=south west}]
        	\addplot[color=blue,mark=x] coordinates {
        		(256, 93.62)
        		(4096, 93.36)
        		(10240, 93.19)
        % 		(25600, 92.3)
        	};
        	\addplot[color=red, mark=x] coordinates {
        		(256, 94.15)
        		(4096, 94.23)
        		(10240, 94.03)
        % 		(25600, 93.47)
        	};
        	legend style={at={(0.03,0.5)},anchor=west}
        \legend{Adam, Adam-LAWN}
        \end{semilogxaxis}%
    \end{tikzpicture}%
\label{fig:adam-cifar-10}}
\quad
\subfigure[CIFAR-100]{%
    \pgfplotsset{every axis/.append style={
        font=\fontsize{16}{16}\selectfont,
        line width=2pt,
        mark size=2.5pt,
        tick style={line width=0.8pt}
    }}
    \begin{tikzpicture}[scale=0.35]
    	\begin{semilogxaxis}[
		    xlabel=Batch size,ylabel=Test Accuracy,
		    ymin=67, ymax=74,
		    ymajorgrids,
		    xmajorgrids,
		    legend style={at={(0.03,0.03)},
		    anchor=south west}]
        	\addplot[color=blue,mark=x] coordinates {
        		(256, 70.84)
        		(4096, 68.91)
        		(10240, 68.61)
        % 		(25600, 69.17)
        	};
        	\addplot[color=red, mark=x] coordinates {
        		(256, 72.99)
        		(4096, 73.12)
        		(10240, 72.97)
        % 		(25600, 71.76)
        	};
        	legend style={at={(0.03,0.5)},anchor=west}
        \legend{Adam, Adam-LAWN}
        \end{semilogxaxis}%
    \end{tikzpicture}%
\label{fig:adam-cifar-100}}
\quad
\subfigure[MovieLens-1M]{%
    \pgfplotsset{every axis/.append style={
        font=\fontsize{16}{16}\selectfont,
        line width=2pt,
        mark size=2.5pt,
        tick style={line width=0.8pt}
    }}
    \begin{tikzpicture}[scale=0.35]
    	\begin{semilogxaxis}[
		    xlabel=Batch size,ylabel=Test HR@10,
		    ymin=68, ymax=71,
		    ymajorgrids,
		    xmajorgrids,
		    legend style={at={(0.03,0.03)},
		    anchor=south west}]
        % 	\addplot[color=blue,mark=x, dashed] coordinates {
        % 		(1000, 0.694)
        % 		(10000, 0.691)
        % 		(100000, 0.672)
        % 		(400000, 0.659)
        % 	};
        	\addplot[color=blue, mark=x] coordinates {
        		(1000, 69.87)
        		(10000, 70.12)
        		(100000, 69.28)
        		(400000, 68.99)
        	};
        	\addplot[color=red,mark=x] coordinates {
        		(1000, 70.80)
        		(10000, 70.41)
        		(100000, 70.77)
        		(400000, 70.66)
        	};
        	legend style={at={(0.03,0.5)},anchor=west}
        \legend{Adam, Adam-LAWN}
        \end{semilogxaxis}%
    \end{tikzpicture}%
\label{fig:adam-ml1m}}
\quad
\subfigure[ImageNet]{%
    \pgfplotsset{every axis/.append style={
        font=\fontsize{16}{16}\selectfont,
        line width=2pt,
        mark size=2.5pt,
        tick style={line width=0.8pt}
    }}
    \begin{tikzpicture}[scale=0.35]
    	\begin{semilogxaxis}[
		    xlabel=Batch size,ylabel=Test Accuracy,
		    ymin=68, ymax=77,
		    ymajorgrids,
		    xmajorgrids,
		    legend style={at={(0.03,0.03)},
		    anchor=south west}]
        	\addplot[color=blue,mark=x] coordinates {
        		(256, 71.16)
        		(16000, 70.84)
        % 		(10240, 68.61)
        % 		(25600, 69.17)
        	};
        	\addplot[color=red, mark=x] coordinates {
        		(256, 76.18)
        		(16000, 76.07)
        % 		(10240, 72.97)
        % 		(25600, 71.76)
        	};
        	legend style={at={(0.03,0.5)},anchor=west}
        \legend{Adam, Adam-LAWN}
        \end{semilogxaxis}%
    \end{tikzpicture}%
\label{fig:adam-imagenet}}
\vspace{-1mm}
\caption{Adam-LAWN vs. Adam (weight decay comprehensively tuned) for a variety of datasets. Adam-LAWN causes little to no drop in generalization performance with increasing batch size.}
\label{fig:adamlawn_bs}
\vspace{-5mm}
\end{figure*}
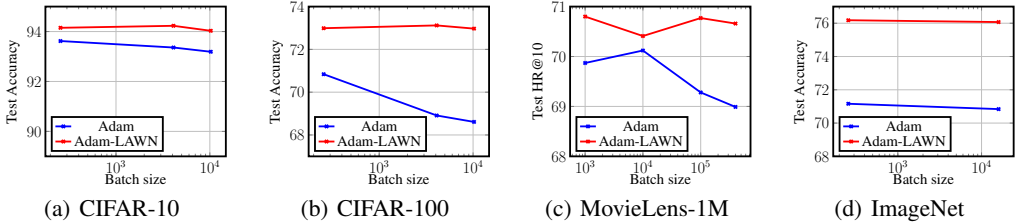

\textbf{Effect of batch size.} 
%Compared to their vanilla counterparts, 
LAWN variants cause more graceful degradation of performance with batch size, as compared to base variants. Adam-LAWN causes almost no degradation in generalization performance even at batch size 10k (see Figures~\ref{fig:adam-cifar-10}, \ref{fig:adam-cifar-100}). 

%\keerthi{Point to appendix for $E_{free}$ values}
\textbf{Effect of $\mathbf{E_{free}}$.} We observed that switching early to LAWN mode (i.e. fixing $E_{free}$ to less than 10 epochs) usually works well for generalization. See Appendix \ref{app:experiments} for details. This is consistent with our hypothesis that constrained training should kick in before loss flattening sets in. 

%%%%%%%%%%%%%%%%%%%%%%%%%%%
%%%%%%%%%%%%%%%%%%%%%%%%%%%
% BEGIN CIFAR ACCURACY vs BATCH_SIZE PLOT
% END CIFAR ACCURACY vs BATCH_SIZE PLOT
%%%%%%%%%%%%%%%%%%%%%%%%%%%
%%%%%%%%%%%%%%%%%%%%%%%%%%%
% \begin{wrapfigure}{R}{4.5cm}
% \centering
% \vspace{-0.5cm}

%     \includegraphics[width=0.36\columnwidth]{images/constrainedstepofsgd.jpg}
%     \vspace{-0.2cm}
%     \caption{\textit{A constrained training step of SGD-LAWN.}}
%     \label{fig:sgdlawn}
% \vskip -0.1in
% \end{wrapfigure}

\newcommand\Tstrut{\rule{0pt}{2.6ex}}         % = `top' strut
\newcommand\Bstrut{\rule[-0.9ex]{0pt}{0pt}}   % = `bottom' strut

\begin{wraptable}{R}{7cm}
% \begin{table*}[!ht]
    \vspace{-0.5cm}

\centering
\resizebox{0.5\textwidth}{!}{%
\begin{tabular}{l|lll|lll}
\hline
\multirow{2}{*}{\textbf{Method}}  &
  \multicolumn{3}{c}{\textbf{CIFAR-10}} &
  \multicolumn{3}{c}{\textbf{CIFAR-100}} \Tstrut\\
  %\hline
    &
  \multicolumn{1}{c}{256} &
  \multicolumn{1}{c}{4k} &
  \multicolumn{1}{c}{10k} &
  \multicolumn{1}{c}{256} &
  \multicolumn{1}{c}{4k} &
  \multicolumn{1}{c}{10k} \Bstrut\\
  \hline
SGD    & \textbf{93.99}&  93.48&  92.99&   73.49&  71.68&  71.07 \Tstrut\\
SGD-L  & 93.96&  \textbf{93.50}&  \textbf{93.43}&   \textbf{73.67}&  \textbf{72.71}& \textbf{71.80} \Bstrut\\
\hline
Adam   & 93.48&  92.93&  92.63&  70.84&  68.91&  68.61  \Tstrut\\
Adam-L & \textbf{93.91}&  \textbf{93.74}&  \textbf{93.84}&  \textbf{72.99}&  \textbf{73.12}&  \textbf{72.97} \Bstrut\\
\hline
LAMB   &  \textbf{93.76}&  \textbf{93.27}&  92.91&  \textbf{71.29}&  69.39&  67.76 \Tstrut\\
LAMB-L &  93.67&  93.22&  \textbf{92.92}&   71.25&  \textbf{69.68}&  \textbf{69.16} \Bstrut\\
\hline
\end{tabular}
}
\caption{Test Acc. on CIFAR-10 and CIFAR-100. Standard error in the the range $[0.1,0.45]$. Details in Appendix~\ref{app:experiments}.}
\label{tab:img_result}
% \end{table*}
\vspace{-0.5cm}
\end{wraptable}

\subsection{Recommendation Systems}

% Introduction to recommendation systems experiments
We conducted experiments on the MovieLens-100k, MovieLens-1M and Pinterest datasets using the popular neural collaborative filtering (NCF) technique \citep{he2017neural}. The datasets contain ratings provided to various items by users.  The model is a 3-layer MLP, the input being user and item embeddings. We use the \textit{hit ratio@10} metric (expressed as \%), where we reward the model for ranking a test item in the top 10 of 100 randomly sampled items that a user has not interacted with in the past. We trained for 300 epochs for the two smallest batch sizes, and 500 epochs for the two biggest batch sizes for each dataset. Details about the datasets, pre-processing, model and evaluation can be found in Appendix \ref{app:experiments}. 
% To test the efficacy of the LAWN technique on recommendation tasks, we consider the task of item prediction using the popular neural collaborative filtering (NCF) technique \cite{he2017neural} on three datasets: MovieLens-100k, MovieLens-1M \cite{harper2015movielens} and Pinterest \cite{geng2015learning}. The datasets contain ratings provided to various items by users, the items being movies for the MovieLens datasets and pins for the Pinterest dataset. More information about the datasets can be found in Table \ref{tab:recodatasets}. Dataset pre-processing information can be found in Appendix \ref{sec:expdetails}
A summary of the performance of the aforementioned optimizers on all 3 datasets can be found in Table \ref{tab:rec_result}. LAWN-based optimizers consistently outperform their base variants. SGD and SGD-LAWN failed to generalize well at large batch sizes and this requires further investigation.

%%%%%%%%%%%%%%%%%%%%%%%%%%%
%%%%%%%%%%%%%%%%%%%%%%%%%%%
% BEGIN MAIN TABLE FOR RECO SYSTEMS
\begin{table*}[!ht]
\vspace{-3mm}
\centering
\resizebox{0.95\textwidth}{!}{%
\begin{tabular}{c|cccc|cccc|cccc}
\hline
\multirow{2}{*}{\textbf{Method}} &
  \multicolumn{4}{c}{\textbf{MovieLens-100k}} &
  \multicolumn{4}{c}{\textbf{MovieLens-1M}} &
  \multicolumn{4}{c}{\textbf{Pinterest}} \Tstrut\\
&
  \multicolumn{1}{c}{1k} &
  \multicolumn{1}{c}{10k} &
  \multicolumn{1}{c}{100k} &
  \multicolumn{1}{c}{400k} &
  \multicolumn{1}{c}{1k} &
  \multicolumn{1}{c}{10k} &
  \multicolumn{1}{c}{100k} &
  \multicolumn{1}{c}{1M} &
  \multicolumn{1}{c}{1k} &
  \multicolumn{1}{c}{10k} &
  \multicolumn{1}{c}{100k} &
  \multicolumn{1}{c}{1M} \Bstrut\\
  \hline
SGD    &66.33  &65.58  & Fail  & Fail  &70.91  &69.31  & Fail  & Fail  &\textbf{86.62}  &85.57  & Fail  &  Fail  \Tstrut\\
SGD-L  &\textbf{66.91}  &\textbf{66.49}  & Fail  & Fail  &70.91  &\textbf{70.41}  & Fail  & Fail  &86.61  &\textbf{85.99}  & Fail  & Fail  \Bstrut\\
\hline
Adam   &66.01  &66.03  &63.20  &63.98  &69.87  &70.12  &69.28  &68.99  &\textbf{87.27}  &85.97  &85.81  &85.30  \Tstrut\\
Adam-L &\textbf{66.81} &\textbf{66.91}  &\textbf{66.24}  &\textbf{66.14}  &\textbf{70.80}  &\textbf{70.41}  &\textbf{70.77}  &\textbf{70.66}  &86.85  &\textbf{86.61}  &\textbf{86.04}  &\textbf{86.06}  \Bstrut\\
\hline
LAMB   &65.45  &65.34  &64.23  &62.57  &69.91  &69.77  &69.44  &68.95  &86.63  &85.91  &85.80  &85.65  \Tstrut\\
LAMB-L &\textbf{66.56}  &\textbf{66.54}  &\textbf{66.52}  &\textbf{66.14}  &\textbf{70.86}  &\textbf{70.86} &\textbf{70.68} &\textbf{70.34}  &\textbf{86.83}  &\textbf{86.25}  &\textbf{85.99}  &\textbf{86.07}  \Bstrut\\
\hline
\end{tabular}%
}
\caption{Test HR@10 on MovieLens and Pinterest recommendations. 
%SGD and SGD-L failed (\textit{F}) to generalize well at large batch sizes and requires further investigation. 
Standard error is in the range $[0.15, 0.25]$, details in Appendix~\ref{app:experiments}.}
\vspace{-3mm}
\label{tab:rec_result}
\end{table*}

% END MAIN TABLE FOR RECO SYSTEMS
%%%%%%%%%%%%%%%%%%%%%%%%%%%
%%%%%%%%%%%%%%%%%%%%%%%%%%%

\textbf{Weight decay vs. LAWN.} Weight decay was used and tuned for all the base optimizers since it arrests the uncontrolled growth of network weights, helping avoid of loss of adaptivity. The LAWN variants do not use weight decay but still outperform the base variants.

\textbf{Effect of batch size.} LAWN variants of Adam and LAMB scale to very large batch sizes (1 million for MovieLens-1M, 400k for MovieLens-100k) without any appreciable loss in accuracy. Both SGD and SGD-LAWN could only scale to batch size 10k. Adam-LAWN's strong scalability with batch size is consistent with results obtained from the CIFAR experiments.

\textbf{Effect of $\mathbf{E_{free}}$.} Similar to the results of the CIFAR experiments, fixing $E_{free}$ to a small value works well for LAWN. Details are in Appendix \ref{app:experiments}.

\vspace{-0.1cm}
\subsection{Image Classification for ImageNet}

As compared to CIFAR, the ImageNet classification problem~\citep{krizhevsky2012imagenet} is more representative of real world classification problems. We used a variant of the popular ResNet50~\citep{he2015deep} model as the classifier. We considered a small (256) and a large (16k) batch size for this experiment, and fixed training budget to be 90 epochs. % For vanilla variants of optimizers, we tuned $E_{warmup}$, learning rate schedule and weight decay. For LAWN variants, we tuned $E_{free}$, $E_{warmup}$ and learning rate schedule. More details about tuning can be found in Appendix \ref{sec:expdetails}.

\textbf{Results for batch size 256.} 
Overall results can be found in Table \ref{tab:res-imagenet} (see \S\ref{sec:intro}). SGD, used in conjunction with momentum and weight decay, has long been the optimizer of choice for image classification. We retain the tuned value for weight decay of the base SGD optimizer for the free phase of SGD-LAWN experiments. SGD-LAWN marginally outperforms SGD. 

Adam is well known to perform worse than SGD for image classification tasks~\citep{wilson2017marginal}. For our experiment, we tuned the learning rate and could only get an accuracy of 71.16\%. In comparison, Adam-LAWN achieves an accuracy of more than 76\%, marginally surpassing the performance of SGD-LAWN and SGD.

We found it difficult to reproduce ImageNet results using the LAMB algorithm. We made minor modifications (details in Appendix \ref{app:experiments}) to the original algorithm to make it more stable, and call the resultant algorithm LAMB+. LAMB-LAWN (the LAWN version of the unmodified LAMB) comprehensively outperforms LAMB+ for batch size 256 by achieving an accuracy close to 76.5\%. 

\textbf{Results for batch size 16k.} 
For the large batch size of 16k, we noticed that LAWN retains strong generalization performance. Both Adam-LAWN and LAMB-LAWN achieve very high accuracy, with Adam-LAWN retaining its performance at such a large batch size by crossing the 76\% test accuracy mark. This is with only additonally tuning for the LAWN variants $E_{free}$ and $E_{warmup}$.

\vspace{-0.3cm}
\section{Conclusion}
\label{sec:conclusion}
\vspace{-0.2cm}
%\keerthi{Aman to update}

In this paper we develop LAWN as a simple and powerful method of modifying deep net training with a base optimizer to improve weight adaptivity and lead to improved generalization. Switching from free to weight norm constrained training at an appropriate point is a key element of the method. We study the performance of the LAWN technique on a variety of tasks, optimizers and batch sizes, demonstrating its efficacy. Tremendous overall enhancement of Adam and the improvement of all base optimizers at large batch sizes using LAWN are important highlights.

%We present a novel technique to extend the capabilities of existing optimizers towards improved generalization performance and scaling for a well-studied class of deep neural networks. We study the performance of the LAWN technique on a wide variety of optimizers and tasks, demonstrating it's efficacy. \textcolor{red}{The key contribution of this paper is that weight norms need to be controlled to avoid crowding with small values, and loss flattening with large values. This is cleanly achieved by LAWN and demonstrated powerfully through experiments.} 

%{\bf Future work.} Let us make a list of things that reviewers would have expected us to do. Later we can combine these in to `future work'.

%\begin{enumerate}
    %\item Any reason why you did not do any textual datasets?
    %\item Can you come up with an automated way of deciding the switch point?
    %\item Is LAWN useful for noisy datasets?
    %\item You talked about moving from ehsn to `All' layers. Can you say how much benefit came from moving to `All'?
%\end{enumerate}

%%%%%%%%%%%%%%%%%%%%%%%%%%%%%%%%%%%%%%%%%%%%%%%%%%%%%%%%%%%%
\bibliographystyle{abbrv}
\bibliography{neurips_2021}
% \appendix

% \section{Appendix}

% Optionally include extra information (complete proofs, additional experiments and plots) in the appendix.
% This section will often be part of the supplemental material.

%%%%%%%%%%%%%%%%%%%%%%%%%%%%%%%%%%%%%%%%%%%%%%%%%%%%%%%%%%%%

%\pagebreak
\newpage

\appendix

\section{Related Work}
\label{sec:relatedwork}

%\keerthi{Add a comment or two along the lines of - most of the key papers have been put in the main paper, we are just putting them all together here.}

The main paper contextually covers all related works at appropriate places. 
%Recent efforts towards optimization for deep neural networks (DNNs) have spawned many different directions of research, so we focus on 
In this section we collect the key papers that are closely related to the LAWN technique and briefly discuss them.

%\keerthi{Add a paragraph about footnote 3 - Weight norm bounding in neural nets is theoretically well-founded for improving generalization [29,3]}

Modern DNNs are largely “over-parameterized”, i.e. the number of model parameters outnumber the training samples. For example, modern convolutional neural networks~\citep{he2015deep, krizhevsky2012imagenet} have tens to hundreds of millions of parameters, and provide state-of-the-art classification performance on datasets with a few million samples~\citep{deng2009imagenet}. 
%Coupled with the use of exponential-type loss functions like logistic loss and cross entropy, over-parameterized DNNs can easily achieve perfect or near-perfect training accuracy by classifying the training data perfectly. During unregularized model training, networks weights are pushed to very high values in order to drive the loss down to small values. With exponential-type loss functions, small loss values yield small gradients, which makes it difficult to update the network weights. This can make it difficult for models to generalize better.\keerthi{This paragraph should go to the Introduction section.}

% \keerthi{Write suitable paragraph, discussing references 2-4 and 5-7 from reviewer 1 of the FixNorm paper.}

The issue of loss flattening and the resulting loss of adaptivity is highlighted in~\citep{szegedy2016rethinking}.
Several well-known techniques are used to mitigate this issue.
%\textit{loss flattening} and \textit{loss of adaptivity} 
%\rohan{cite} 
%of the network. 
$\ell_2$ regularization is often used in conjunction with optimizers like SGD, heavy-ball momentum \cite{qian1999momentum} and adaptive optimizers \cite{kingma2017adam, zeiler2012adadelta, duchi2011adaptive},  among others. Recently, weight decay \cite{loshchilov2019decoupled, hanson1988comparing}, which is not the same as $\ell_2$ regularization for adaptive optimizers, has become popular to reign-in network weights and prevent the model from becoming overconfident on training samples. %Another effective technique is 
Other techniques are: (a) label smoothing regularization \cite{szegedy2016rethinking}, which makes the model less confident about predictions by changing the ground-truth label distribution; and (b) flooding~\cite{ishida2020we} which tries to keep the aggregate training loss to be around a specified small value.  
%Other recent works that directly or indirectly discuss preventing the weights from increasing to very large values include \cite{salimans2016weight, hoffer2018norm, ishida2020we}. 

From a theory perspective, weight norm bounding has been shown to be useful for improving generalization~\cite{NeyshaburTS15, BartlettFT17}.
Salimans and Kingma~\cite{salimans2016weight} uses weight normalization as a transformation; but, by also keeping the scale component, it ends up allowing logits to grow large.
%\rohan{also do norm clipping? but - add more context} 
%using both the radial and lateral component of the gradient, which is wasteful.  
Hoffer et al~\cite{hoffer2018norm} discusses keeping the norms of the parameters fixed, but it 
%\anika{would require (I think "requires" make it seem like Hoffer already suggested a LAWN-like method)} 
would require a LAWN-like method to work effectively.  

A perplexing property of SGD-like optimization methods is their ability to provide strong generalization performance. It has been observed that using small batch sizes makes it easier to extract better generalization performance from networks~\citep{keskar2016large}. The noise injected into the optimization procedure by stochastic mini-batch gradients and the learning rate help improve generalization performance. This makes it difficult to use large batch sizes for optimization, where the noise component of the gradient diminishes with increasing batch sizes. Large batch sizes are interesting because they can help speed up the training process by leveraging multiple machines. Poor generalization performance for large batch sizes is attributed to them stalling around "sharp" minimizers~\citep{keskar2016large}.
%, and "sharp" minimizers result in worse generalization. 
Goyal et al~\citep{goyal2018accurate} scale ImageNet 
%~\citep{deng2009imagenet} 
training to batch sizes of 8k without obvious loss in generalization performance, by carefully tuning parameters like learning rate and batch normalization. Other recent efforts to train large-batch models include ~\citep{hoffer2017train, you2017large, you2019large, shallue2018measuring, you2020large}. It is important to note that for a majority of the cited works, large batch gains do not necessarily hold across tasks or datasets.

Recent theory has shown implicit bias of optimizers to maximize the (normalized) $\ell_2$-margin %(which has links to robustness) 
of linear models ~\citep{soudry2018implicit, gunasekar2018implicit} and fully homogeneous neural networks ~\citep{lyu2020gradient, Poggio2019, Poggio2020, Wang2020}. LAWN and $\ell_2$ regularization approximate this implicit bias property. 
%(we define homogeneous neural networks later in the paper) under assumptions like perfect separability of training data. 

%TODO expand on this. TODO add details on escaping sub-optimal minima.

%TODO add details on direct margin maximization. Cite this paper - https://arxiv.org/pdf/1806.00730.pdf 

\section{Loss Flattening}
\label{sec:lossflattening}

% \keerthi{Add an introductory line about the problem setting. Also mention that similar things will happen for a multi-class setting. }

\begin{wrapfigure}{R}{6cm}
\centering
\vspace{-1.8cm}

\includegraphics[width=0.4\textwidth]{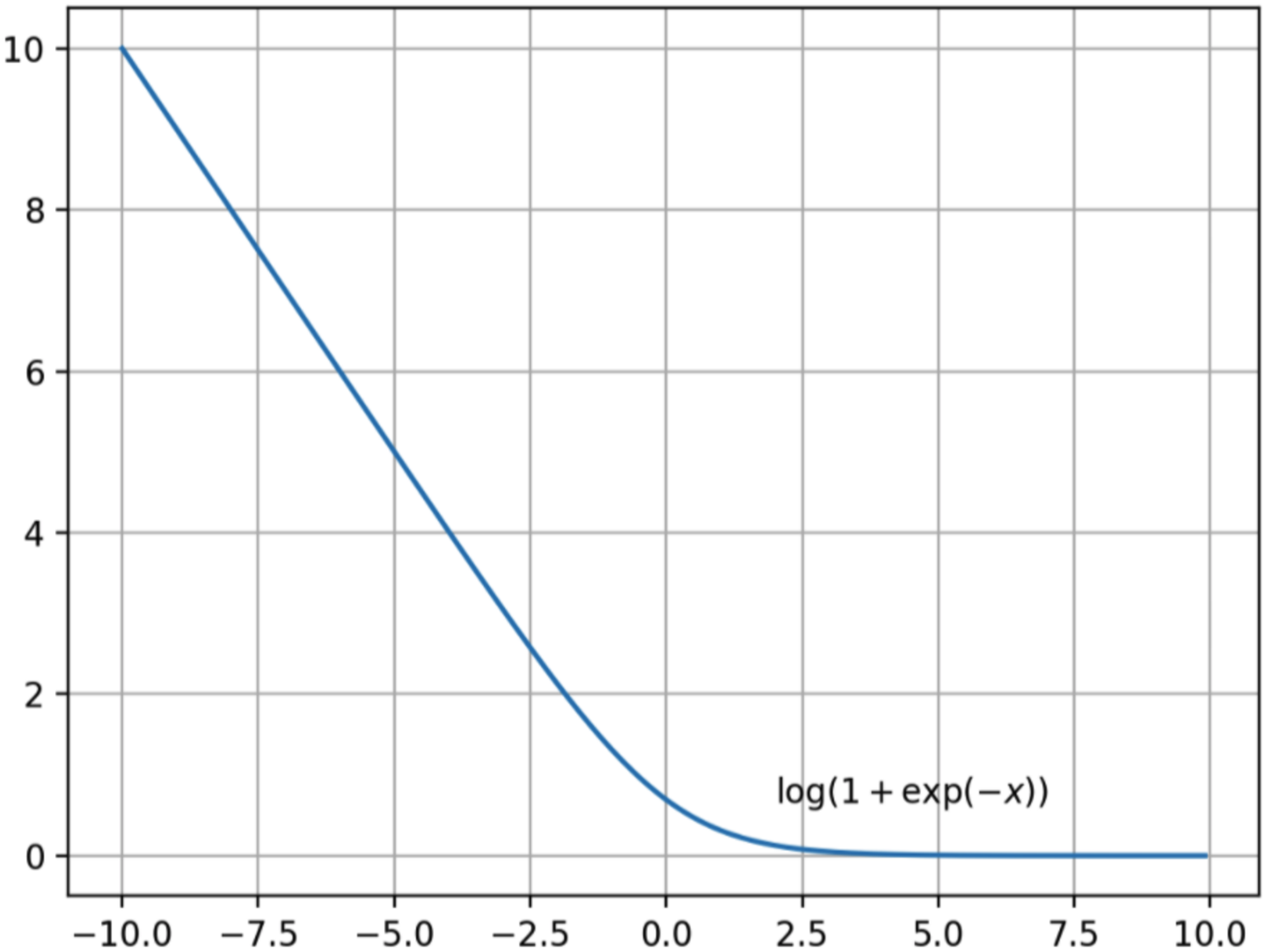}
\caption{Logistic loss becomes small and flat if the logit crosses a threshold of 2.5 or more. Training becomes less adaptive when the losses of most examples fall in the flat zone. 
%making it difficult for training to adaptive.
}
\label{fig:loss_flattening}
% \vspace{-1.2cm}
\end{wrapfigure}

Consider a binary classification problem. Figure \ref{fig:loss_flattening} depicts the logistic loss as a function of logit. Notice that loss becomes extremely small once the logit value crosses a certain threshold (say 2.5). A small value of loss yields small gradients, making it difficult for neural network training to be adaptive. We define this as \textit{loss flattening}. Multiclass classification is slightly more complex. The difference between the score of the target class and the largest score of the remaining classes plays the role of the logit and, when it becomes large enough, loss flattening occurs.

    \begin{algorithm}[H]
        \caption{ResetState($m$, $v$)}
        \label{algo:project}
        %\DontPrintSemicolon
        \begin{algorithmic}[1]
        \State \textbf{Input}: First moment $m$ and second moment $v$
        \State \textbf{Output}: Updated $m$ and $v$
        \State $m$ = 0
        \State $v$ = 0
        \end{algorithmic}
    \end{algorithm}

      \begin{algorithm}[H]
        \caption{Project($g^{\ell}$, $w^{\ell}$, $c^{\ell}$, $k$, $t$)}
        \label{algo:project}
        %\DontPrintSemicolon
        \begin{algorithmic}[1]
        \State \textbf{Input}: Gradient or gradient-like update $g^\ell$, weights $w_\ell$ and norm $c_\ell$ for layer $\ell$, LAWN switch step $k$, current step $t$.
        \State \textbf{Output}: Projected gradient $h^\ell$.
      \If{$t$ < $k$}
        \State $h^\ell$ = $g^\ell$
      \Else
        \State $h^\ell = g^\ell - \frac{w^\ell.g^\ell}{(c^\ell)^2} w^\ell$
      \EndIf
        \end{algorithmic}
    \end{algorithm}

    \begin{algorithm}[h]
        \caption{NormalizeWeights($w_{t}^{\ell}$, $c^{\ell}$, $k$, $t$)}
        \label{algo:normalize}
        %\DontPrintSemicolon
        \begin{algorithmic}[1]
        \State \textbf{Input}: Weights $w_{t}^{\ell}$ and norm $c^{\ell}$ for layer $\ell$, LAWN switch step $k$, current step $t$.
        \State \textbf{Output}: Normalized weights $nw_{t}^{\ell}$.
      \If{$t$ < $k$}
        \State $nw_{t}^{\ell}$ = $w_{t}^{\ell}$
      \Else
        \State $nw_{t}^{\ell}$ = $\frac{c^{\ell}}{||w_{t}^{\ell}||} w_{t}^{\ell}$
      \EndIf
        \end{algorithmic}
    \end{algorithm}
    
    \begin{algorithm}[h]
        \caption{UpdateWeights($g_{t}^{\ell}$, $w_{t-1}^{\ell}$, $\lambda$, $k$, $t$)}
        \label{algo:normalize}
        %\DontPrintSemicolon
        \begin{algorithmic}[1]
        \State \textbf{Input}: Gradient or gradient-like update $g_{t}^{\ell}$ and weights $w_{t-1}^{\ell}$ for layer $\ell$, weight decay factor $\lambda$, LAWN switch step $k$, current step $t$.
        \State \textbf{Output}: Updated weights $\hat{g}_{t}^{\ell}$.
      \If{$t$ < $k$}
        \State $\hat{g}_{t}^{\ell}$ = $g_t^{\ell} + \lambda w_{t-1}^{\ell}$
       \Else
        \State $\hat{g}_{t}^{\ell}$ = $g_t^{\ell}$
      \EndIf
        \end{algorithmic}
    \end{algorithm}

\section{Detailed LAWN algorithms}
\label{app:lawnalgos}

\begin{wrapfigure}{R}{5cm}
\centering
\vspace{-1.5cm}

    \includegraphics[width=0.31\columnwidth]{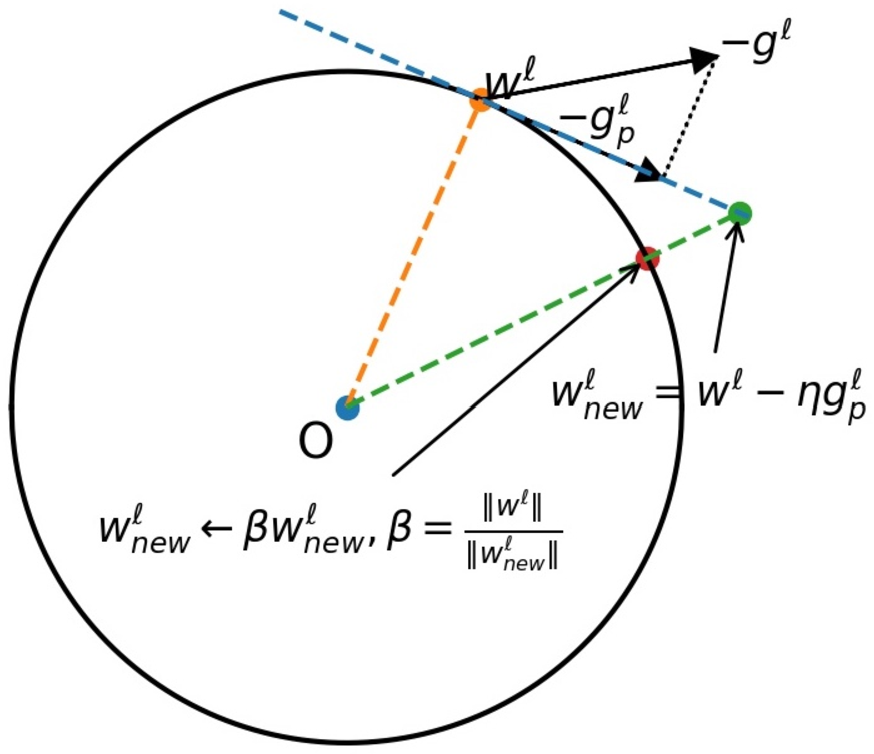}
    \vspace{-0.2cm}
    % \caption{\textit{A constrained training step of SGD-LAWN.}}
    \label{fig:sgdlawn}
% \vskip -0.1in
\end{wrapfigure}

Consider the optimization of a network consisting of L layers. We apply LAWN to a given layer $\ell$ by switching to constrained training at a suitable point during the optimization group. An example of $\ell$ is the combination of the weight and bias terms for a feed forward layer, where we consider the norms of both the weight and bias as a whole after switching to constrained training.

The figure on the right shows a constrained training step of SGD-LAWN for updating $w^{\ell}$, the weight subvector of a generic group. The weight vector is updated using the projected gradient. The updated weight vector is then normalized in order to ensure that it has the same euclidean norm as the previous iterate.

We now provide detailed algorithms for Adam-LAWN and LAMB-LAWN. For each LAWN version, we do a side-by-side comparison with its regular version. Differences are highlighted in pink.

\begin{minipage}{\textwidth}
  \begin{minipage}[t]{0.4\textwidth}
    \begin{algorithm}[H]
    \caption{Adam With Weight Decay}
    \label{algo:adamw}
    %\DontPrintSemicolon
    \begin{algorithmic}[1]
    
    \State \textbf{Input}: Training set $P$, Network layers $L$, $w_0 \in R^d$, learning rates $\{{\eta_t}\}_{t=0}^{T}$, parameters $\epsilon > 0$ and $0 < \beta_1, \beta_2 < 1$, weight decay factor $\lambda \in R$.
    \State Initialize variables $m_0 = 0, v_0 = 0$.
    
    \For {t  in 1...T}
        \State Draw batch $S_t$ from $P$
        \State $g_t = $ ComputeGrad($w_{t-1}$, $S_t$)
        \State $m_t = \beta_1 m_{t-1} + (1-\beta_1)g_t$
        \State $v_t = \beta_2 v_{t-1} + (1-\beta_2)g_t^2$
        \State $\hat{m}_t = \frac{m_t}{1-\beta_1^t}, \hat{v}_t = \frac{v_t}{1-\beta_2^t}$
        \State Compute $r_t = \frac{\hat{m}_t}{\sqrt{\hat{v}_t} + \epsilon}$
        \State Compute $\hat{r}_t^{\ell} = r_t^{\ell} + \lambda w_{t-1}^{\ell}$ 
        \State $w_{t}^{\ell} = w_{t-1}^{\ell} - \eta_t\hat{r}_t^{\ell}$   $ \forall \ell$
    \EndFor
    \end{algorithmic}
    \end{algorithm}
  \end{minipage}
  \hfill
  \begin{minipage}[t]{0.57\textwidth}
    \begin{algorithm}[H]
    \caption{Adam-LAWN}
    \label{algo:adamlawn}
    %\DontPrintSemicolon
    \begin{algorithmic}[1]
    
    \State \textbf{Input}: Training set $P$, Network layers $L$, $w_0 \in R^d$, learning rates $\{{\eta_t}\}_{t=0}^{T}$, parameters $\epsilon > 0$ and $0 < \beta_1, \beta_2 < 1$, weight decay factor $\lambda \in R$, lawn switch step $1 <= k <= T$.
    \State Initialize variables $m_0 = 0, v_0 = 0$.
    
    \For {t  in 1...T}
        \State Draw batch $S_t$ from $P$
        \If{\colorbox{pink!30}{$t$ == $k$}}
        \State {\colorbox{pink!30}{ResetState($m_{t-1}, v_{t-1}$), $c^{\ell} = ||w_{t-1}^{\ell}||$ $ \forall \ell$}}
        \EndIf
        \State $g_t = $ ComputeGrad($w_{t-1}$, $S_t$)
        \State \colorbox{pink!30}{$g_{t}^{\ell} \leftarrow $  Project(
 $g_{t}^{\ell}$, $w_{t-1}^{\ell}$, $c^{\ell}$, $k$, $t$) $ \forall \ell$}
        \State $m_t = \beta_1 m_{t-1} + (1-\beta_1)g_t$
        \State $v_t = \beta_2 v_{t-1} + (1-\beta_2)g_t^2$
        \State $\hat{m}_t = \frac{m_t}{1-\beta_1^t}, \hat{v}_t = \frac{v_t}{1-\beta_2^t}$
        \State Compute \colorbox{pink!30}{$r_t^\ell = $ Project($\frac{\hat{m}_t^l}{\sqrt{\hat{v}_t^l} + \epsilon}$, $w_{t-1}^{\ell}$, $k$, $t$) $ \forall \ell$}
        \State Compute \colorbox{pink!30}{$\hat{r}_t^{\ell} = $ UpdateWeights $(r_t^{\ell} , w_{t-1}^{\ell}, \lambda, k, t)$ $ \forall \ell$}
        \State $w_{t}^{\ell} = w_{t-1}^{\ell} - \eta_t\hat{r}_t^{\ell}$ $ \forall \ell$
        \State \colorbox{pink!30}{NormalizeWeights($w_{t}^{\ell}$, $c^{\ell}$, $k$, $t$)  $ \forall \ell$}
    \EndFor
    \end{algorithmic}
    \end{algorithm}
  \end{minipage}
\end{minipage}

\begin{minipage}{\textwidth}
  \begin{minipage}[t]{0.44\textwidth}
    \begin{algorithm}[H]
    \caption{LAMB}
    \label{algo:lamb}
    %\DontPrintSemicolon
    \begin{algorithmic}[1]
    
    \State \textbf{Input}: Training set $P$, Network layers $L$, $w_0 \in R^d$, learning rates $\{{\eta_t}\}_{t=0}^{T}$, parameters $\epsilon > 0$ and $0 < \beta_1, \beta_2 < 1$, weight decay factor $\lambda \in R$, scaling function $\phi$.
    \State Initialize variables $m_0 = 0, v_0 = 0$.
    
    \For {t  in 1...T}
        \State Draw batch $S_t$ from $P$
        \State $g_t = $ ComputeGrad($w_{t-1}$, $S_t$)
        \State $m_t = \beta_1 m_{t-1} + (1-\beta_1)g_t$
        \State $v_t = \beta_2 v_{t-1} + (1-\beta_2)g_t^2$
        \State $\hat{m}_t = \frac{m_t}{1-\beta_1^t}, \hat{v}_t = \frac{v_t}{1-\beta_2^t}$
        \State Compute $r_t = \frac{\hat{m}_t}{\sqrt{\hat{v}_t} + \epsilon}$
        \State Compute $\hat{r}_t^{\ell} = r_t^{\ell} + \lambda w_{t-1}^{\ell}$
        \State $w_{t}^{\ell} = w_{t-1}^{\ell} - \eta_t\frac{\phi(w_{t}^{\ell})}{||\hat{r}_t^{\ell}||}\hat{r}_t^{\ell}$ $ \forall \ell$
        
    \EndFor
    \Statex{{\small LAMB+ corresponds to clipping the trust ratio $\frac{\phi(w_{t}^{\ell})}{||\hat{r}_t^{\ell}||}$ to a max value of 1.0.}}
    \end{algorithmic}
    \end{algorithm}
  \end{minipage}
  \hfill
  \begin{minipage}[t]{0.55\textwidth}
    \begin{algorithm}[H]
    \caption{LAMB-LAWN}
    \label{algo:lamblawn}
    %\DontPrintSemicolon
    \begin{algorithmic}[1]
    
    \State \textbf{Input}: Training set $P$, Network layers $L$, $w_0 \in R^d$, learning rates $\{{\eta_t}\}_{t=0}^{T}$, parameters $\epsilon > 0$ and $0 < \beta_1, \beta_2 < 1$, weight decay factor $\lambda \in R$, lawn switch step $1 <= k <= T$, scaling function $\phi$.
    \State Initialize variables $m_0 = 0, v_0 = 0$.
    
    \For {t  in 1...T}
        \State Draw batch $S_t$ from $P$
        \If{\colorbox{pink!30}{$t$ == $k$}}
        \State {\colorbox{pink!30}{ResetState($m_{t-1}, v_{t-1}$), $c^{\ell} = ||w_{t-1}^{\ell}||$ $ \forall \ell$}}        \EndIf
        \State $g_t = $ ComputeGrad($w_{t-1}$, $S_t$)
        \State \colorbox{pink!30}{$g_{t}^{\ell} \leftarrow $  Project(
 $g_{t}^{\ell}$, $w_{t-1}^{\ell}$, $c^{\ell}$, $k$, $t$) $ \forall \ell$}
        \State $m_t = \beta_1 m_{t-1} + (1-\beta_1)g_t$
        \State $v_t = \beta_2 v_{t-1} + (1-\beta_2)g_t^2$
        \State $\hat{m}_t = \frac{m_t}{1-\beta_1^t}, \hat{v}_t = \frac{v_t}{1-\beta_2^t}$
        \State Compute \colorbox{pink!30}{$r_t^\ell = $ Project($\frac{\hat{m}_t^l}{\sqrt{\hat{v}_t^l} + \epsilon}$, $w_{t-1}^{\ell}$, $k$, $t$) $ \forall \ell$}
        \State Compute \colorbox{pink!30}{$\hat{r}_t^{\ell} = $ UpdateWeights $(r_t^{\ell} , w_{t-1}^{\ell}, \lambda, k, t)$ $ \forall \ell$}
        \State $w_{t}^{\ell} = w_{t-1}^{\ell} - \eta_t\frac{\phi(w_{t}^{\ell})}{||\hat{r}_t^{\ell}||}\hat{r}_t^{\ell}$ $ \forall \ell$
        \State \colorbox{pink!30}{NormalizeWeights($w_{t}^{\ell}$, $c^{\ell}$, $k$, $t$)  $ \forall \ell$}
    \EndFor
    \end{algorithmic}
    \end{algorithm}
  \end{minipage}
\end{minipage}
\section{Experiments}
\label{app:experiments}
\vspace{-0.25cm}

We now describe the detailed experimental setup for all experiments. For Adam, LAMB, Adam-LAWN and LAMB-LAWN, we fixed $\beta_1 = 0.9$ and $\beta_2 = 0.999$. For Adam, we used $\epsilon = 10^{-8}$ and the rest use $10^{-6}$. 
\vspace{-0.25cm}
\subsection{Learning Rate Schedule}
LAWN-based optimizers run unconstrained training for $E_{free}$ epochs, and then switch to constrained training with projected gradients.

We propose using a learning rate schedule that accommodates two stages of warm up (for free and constrained training), followed by learning rate decay in the constrained phase. Both warm up and decay of the learning rate have proved benefitial for generalization performance, even for adaptive optimizers ~\citep{loshchilov2019decoupled, liu2020variance}. For regular optimizers, we use a linear warmup of the learning rate followed by linear decay to zero. Figures~\ref{fig:lrsched_lawn} and~\ref{fig:lrsched_nonlawn} depict the aforementioned schedules.

\begin{minipage}{\textwidth}
  \begin{minipage}[b]{0.48\textwidth}
    \centering
    \includegraphics[width=1.0\columnwidth]{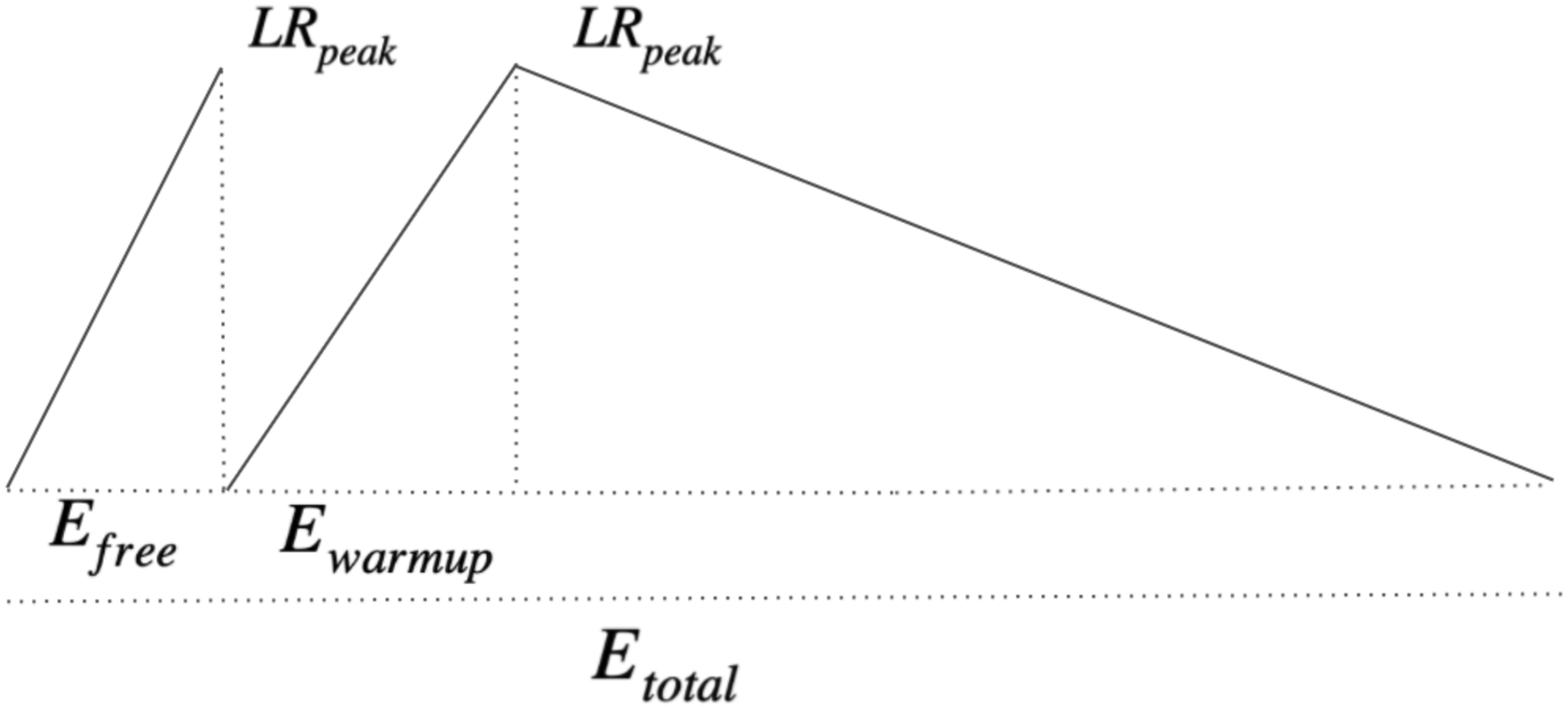}
    \captionof{figure}{\textbf{LAWN learning rate schedule}: The schedule accommodates two linear warmups, followed by a final linear decay over a total of $E_{total}$ epochs.}
    \label{fig:lrsched_lawn}

  \end{minipage}
  \hfill
  \begin{minipage}[b]{0.48\textwidth}
    \centering
    \includegraphics[width=1.0\columnwidth]{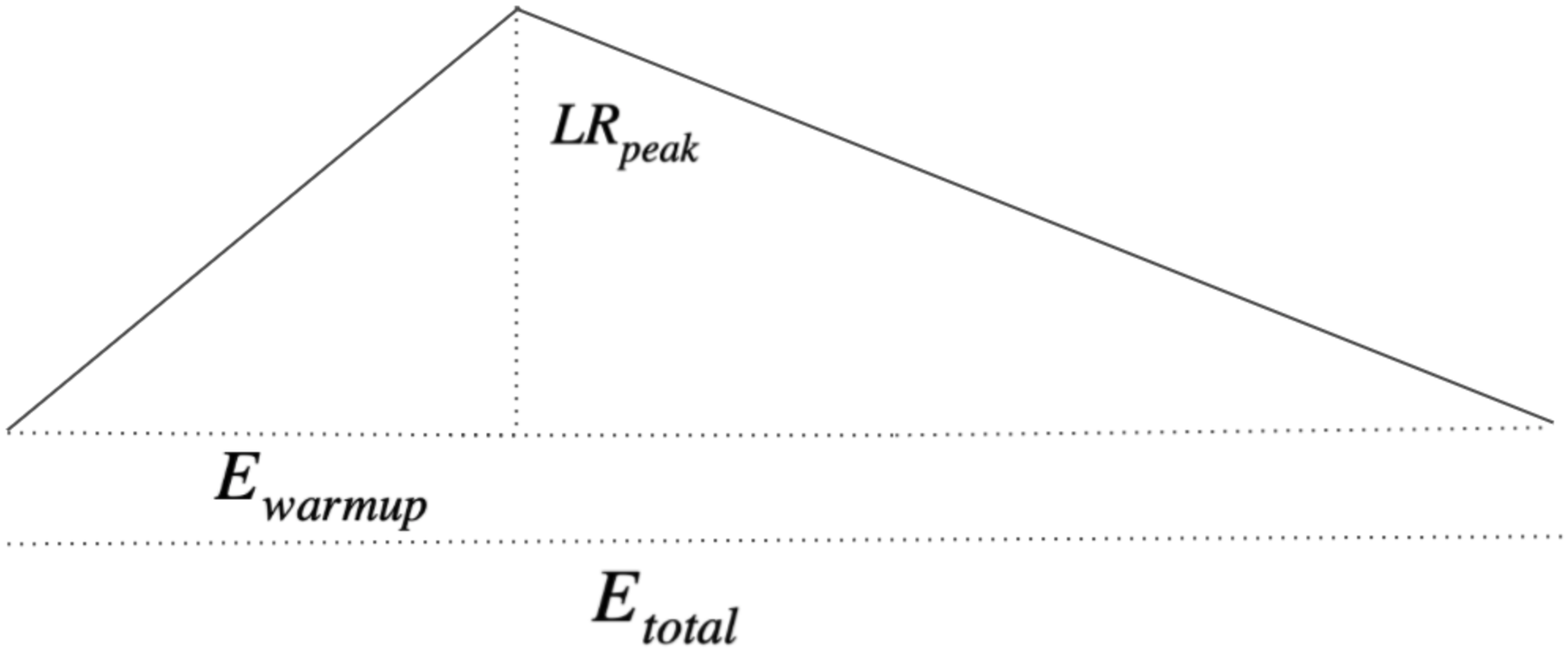}
    \captionof{figure}{\textbf{Regular learning rate schedule}: The schedule accommodates a linear warmup, followed by linear decay over a total of $E_{total}$ epochs.}
    \label{fig:lrsched_nonlawn}
  \end{minipage}
\end{minipage}

% \begin{figure*}[htb!]
%     \includegraphics[width=\textwidth]{images/lawn_lr_schedule_v2.png}
% \caption{Learning rate schedule used by LAWN: The schedule accommodates two linear warmups, followed by a final linear decay over a total of $E_{total}$ epochs.}
% \label{fig:lr_schedule}
% \end{figure*}

\subsection{CIFAR Experimental Setup}

\textbf{Dataset Details} Both CIFAR-10 and CIFAR-100 contain the same set of 50k images for training and 10k images for the test set. CIFAR-10 uses 10 labels, where CIFAR-100 uses 100 labels.

\textbf{Dataset Pre-Processing}
% \lstinputlisting[language=python]{
% tt_cifar10.py
% }
For both CIFAR-10 and CIFAR-100, we employ data augmentation techniques for training. We use $RandomCrop(32, padding=4)$ and $RandomHorizontalFlip$. Additionally, for CIFAR-100, we also use $RandomRotation(15)$. We mean center all the training and test samples.

\textbf{Network Architecture}
We used VGG-19 network with a single fully connected layers for all CIFAR experiments. The network is almost homogeneous except for the batch normalization layer. Link to the code used will be released once the paper is accepted.

%For more details, the reader can refer to this open source code repository \url{https://github.com/weiaicunzai/pytorch-cifar100}. \keerthi{Consider removing the link for the repo.}

\textbf{Hyperparameters}
For each optimizer, we tuned the following parameters: learning rate, $E_{free}$ and $E_{warmup}$.
We use linear learning scheduling where $E_{total}$ is fixed at 300 epochs. $E_{warmup}$ is searched in [5, 20, 40, 60]. $E_{free}$ is searched in [1, 5, 20, 40] for the LAWN methods. Momentum for SGD and SGD-LAWN was fixed at 0.9. Gradient clipping was used for all optimizers with the gradient norm clipped to a maximum value of 1.0.

The peak learning rate $\eta_{peak}$is swept in a two-stage grid search, the grid at the first stage is wide but coarse while the grid at the second stage is narrow but dense near the best point found in the first stage. Six values are searched in the first stage and five points are searched in the second stage. Details can be found in Table~\ref{tab:cifar-lr-se}.

\begin{table*}[!ht]
\centering
%\resizebox{0.75\textwidth}{!}{%
\begin{tabular}{l|l|l|l}
\hline
\textbf{Optimizer} & \textbf{Batch Size} & \textbf{CIFAR-10} & \textbf{CIFAR-100} \Bstrut \Tstrut  \\
\hline
Coarse search &256, 4k, 10k & 1e-5, 1e-4, 1e-3, 0.01, 0.1, 1.0 & 1e-5, 1e-4, 1e-3, 0.01, 0.1, 1.0 \Tstrut  \Bstrut \\
\hline
\multirow{4}{*}{SGD} & 256 & 0.5, 0.7, 0.9, 1.1, 1.3 & 0.20, 0.21, 0.22, 0.23, 0.24 \Tstrut \\ %\cline{2-4}
                     & 4k & 1.5, 1.7, 1.9, 2.1, 2.3 & 0.8, 0.9, 1.0, 1.1, 1.2  \\ %\cline{2-4}
                     & 10k & 1.3, 1.5, 1.7, 1.9, 2.1 & 1.3, 1.4, 1.5, 1.6, 1.7 \Bstrut \\ %\cline{2-4}
\hline
\multirow{4}{*}{SGD-L} & 256 & 0.21, 0.22, 0.23, 0.24, 0.25 & 0.16, 0.17, 0.18, 0.19, 0.20 \Tstrut \\ %\cline{2-4}
                     & 4k & 1.5, 1.6, 1.7, 1.8, 1.9 & 1.2, 1.3, 1.4, 1.5, 1.6  \\ %\cline{2-4}
                     & 10k & 1.0, 1.1, 1.2, 1.3, 1.4& 1.6, 1.7, 1.8, 1.9, 2.0 \Bstrut \\ %\cline{2-4}
\hline
\multirow{4}{*}{Adam} & 256 & [1, 2, 3, 4, 5]$\times$e-3 & [6, 7, 8, 9, 10]$\times$e-3 \Tstrut \\ %\cline{2-4}
                     & 4k & [5, 6, 7, 8, 9]$\times$e-3 & [8, 9, 10, 11, 12]$\times$e-3  \\ %\cline{2-4}
                     & 10k & [7, 8, 9, 10, 11]$\times$e-3 & [8, 9, 10, 11, 12]$\times$e-3 \Bstrut \\ %\cline{2-4}
\hline
\multirow{4}{*}{Adam-L} & 256 & [5, 6, 7, 8, 9]$\times$e-4 & [3, 4, 5, 6, 7]$\times$e-4 \Tstrut \\ %\cline{2-4}
                     & 4k & [4, 5, 6, 7, 8]$\times$e-3 & [1, 2, 3, 4, 5]$\times$e-3  \\ %\cline{2-4}
                     & 10k & [4, 5, 6, 7, 8]$\times$e-3 & [5, 6, 7, 8, 9]$\times$e-3 \Bstrut \\ %\cline{2-4}
\hline
\multirow{4}{*}{LAMB} & 256 & [1, 2, 3, 4, 5]$\times$e-2 & [4, 5, 6, 7, 8]$\times$e-2  \Tstrut \\ %\cline{2-4}
                     & 4k & [1, 2, 3, 4, 5]$\times$e-2 & [5, 7, 9, 11, 13]$\times$e-2  \\ %\cline{2-4}
                     & 10k & [2, 3, 4, 5, 6]$\times$e-2 & [3, 4, 5, 6, 7]$\times$e-2 \Bstrut \\ %\cline{2-4}
\hline
\multirow{4}{*}{LAMB-L} & 256 & [6, 7, 8, 9, 10]$\times$e-3 & [0.5, 1, 2, 3, 4]$\times$e-2 \Tstrut \\ %\cline{2-4}
                     & 4k & [1, 2, 3, 4, 5]$\times$e-2 & [2, 3, 4, 5, 6]$\times$e-2  \\ %\cline{2-4}
                     & 10k & [1, 2, 3, 4, 5]$\times$e-2 & [4, 5, 6, 7, 8]$\times$e-2  \Bstrut \\ %\cline{2-4}
\hline
\end{tabular}
%}
\caption{Peak learning rate values for CIFAR-10 and CIFAR-100.}
\label{tab:cifar-lr-se}
\end{table*}

For non-LAWN methods, weight decay is used. Its value is swept over [1e-5, 1e-4, 1e-3]. Weight decay is not applied in LAWN methods except the free phase of SGD-LAWN. Results with standard error are shown in Table \ref{tab:img_result_se}.
%\begin{wraptable}{R}{7cm}
\begin{table}[!ht]
%\vspace{-0.5cm}

\centering
% \resizebox{0.5\textwidth}{!}{%
\begin{tabular}{l|lll|lll}
\hline
\multirow{2}{*}{\textbf{Method}}  &
  \multicolumn{3}{c}{\textbf{CIFAR-10}} &
  \multicolumn{3}{c}{\textbf{CIFAR-100}} \Tstrut\\
  %\hline
    &
  \multicolumn{1}{c}{256} &
  \multicolumn{1}{c}{4k} &
  \multicolumn{1}{c}{10k} &
  \multicolumn{1}{c}{256} &
  \multicolumn{1}{c}{4k} &
  \multicolumn{1}{c}{10k} \Bstrut\\
  \hline
SGD    & \textbf{93.99} (0.05)&  93.48 (0.09)&  92.99 (0.20)&   73.49 (0.43)&  71.68 (0.26)&  71.07 (0.14) \Tstrut\\
SGD-L  & 93.96 (0.06)&  \textbf{93.50} (0.05)&  \textbf{93.43} (0.12)&   \textbf{73.67} (0.12)&  \textbf{72.71} (0.07)& \textbf{71.80} (0.36) \Bstrut\\
\hline
Adam   & 93.48 (0.11)&  92.93 (0.01)&  92.63 (0.06)&  70.84 (0.23)&  68.91 (0.05)&  68.61 (0.28)  \Tstrut\\
Adam-L & \textbf{93.91} (0.04)&  \textbf{93.74} (0.04)&  \textbf{93.84} (0.08)&  \textbf{72.99} (0.02)&  \textbf{73.12} (0.11)&  \textbf{72.97} (0.10)\Bstrut\\
\hline
LAMB   &  \textbf{93.76} (0.07)&  \textbf{93.27} (0.08)&  92.91 (0.03)&  \textbf{71.29} (0.11)&  69.39 (0.06)&  67.76 (0.22) \Tstrut\\
LAMB-L &  93.67 (0.04)&  93.22 (0.05)&  \textbf{92.92} (0.03)&   71.25 (0.17)&  \textbf{69.68} (0.16)&  \textbf{69.16} (0.09)\Bstrut\\
\hline
\end{tabular}
%}
\vspace{0.25cm}
\caption{Test Acc. on CIFAR-10 and CIFAR-100 with standard error.}
\label{tab:img_result_se}
\end{table}
%\vspace{-0.2cm}
%\end{wraptable}

\begin{figure*}[htbp!]
\centering
\subfigure[CIFAR-10]{%
    \pgfplotsset{every axis/.append style={
        font=\large,
        line width=2pt,
        mark size=2.5pt,
        tick style={line width=0.8pt}
    }}
    \begin{tikzpicture}[scale=0.65]
    	\begin{semilogxaxis}[
		    xlabel=Epochs,ylabel=Test Accuracy,
		    ymin=93.5, ymax=94.5,
		    ymajorgrids,
		    xmajorgrids,
		    legend style={at={(0.03,0.03)},
		    anchor=south west}]
        	\addplot[color=teal,mark=square*] coordinates {
        		(1, 94.1)
        		(5, 94.16)
        		(20, 94.21)
        		(40, 94.0)
        	};
        	\addplot[color=green, mark=square*] coordinates {
        		(1, 94.25)
        		(5, 94.03)
        		(20, 94.14)
        		(40, 94.06)
        	};
        	\addplot[color=olive, mark=square*] 
        	coordinates {
        		(1, 94.17)
        		(5, 93.87)
        		(20, 93.93)
        		(40, 93.77)
        	};
        	legend style={at={(0.03,0.5)},anchor=west}
        \legend{Batch size = 256, Batch size = 4k, Batch size = 10k}
        \end{semilogxaxis}%
    \end{tikzpicture}%
\label{fig:adamlawn-ls-cifar10}}
\quad
\subfigure[MovieLens-100k]{%
    \pgfplotsset{every axis/.append style={
        font=\large,
        line width=2pt,
        mark size=2.5pt,
        tick style={line width=0.8pt}
    }}
    \begin{tikzpicture}[scale=0.65]
    	\begin{semilogxaxis}[
		    xlabel=Epochs,ylabel=Test HR@10,
		    ymin=0.62, ymax=0.68,
		    ymajorgrids,
		    xmajorgrids,
		    legend style={at={(0.03,0.03)},
		    anchor=south west}]
        	\addplot[color=teal,mark=square*] coordinates {
        		(1, 0.668)
        		(5, 0.665)
        		(20, 0.657)
        		(50, 0.651)
        		(100, 0.656)
        	};
        	\addplot[color=green, mark=square*] coordinates {
        		(1, 0.669)
        		(5, 0.656)
        		(20, 0.652)
        		(50, 0.652)
        		(100, 0.645)
        	};
        	\addplot[color=olive, mark=square*] 
        	coordinates {
        		(1, 0.661)
        		(5, 0.659)
        		(20, 0.657)
        		(50, 0.645)
        		(100, 0.638)
        	};

        	legend style={at={(0.03,0.5)},anchor=west}
        \legend{Batch size = 1k, Batch size = 10k, Batch size = 100k}
        \end{semilogxaxis}%
    \end{tikzpicture}%
\label{fig:adam-ml100k-bs}}
\caption{Effect of $E_{free}$ on Adam-LAWN.}
\end{figure*}
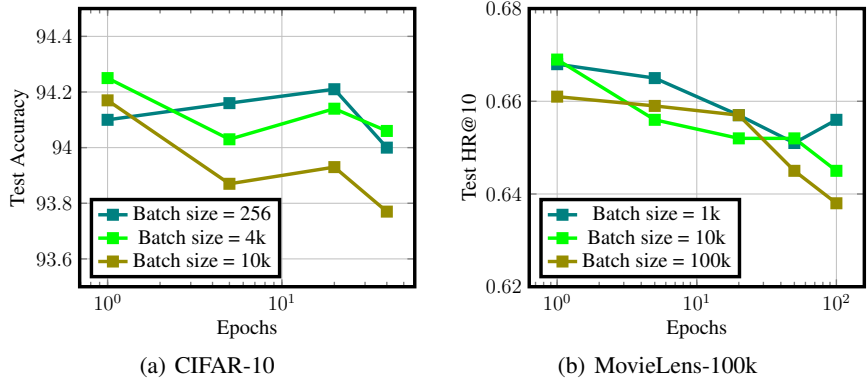

Figure \ref{fig:adamlawn-ls-cifar10} shows the effect of LAWN switch point $E_{free}$ on CIFAR-10 when Adam-LAWN is used. 

%%%%%%%%%%%%%%%%%%%%%%%%%%%
%%%%%%%%%%%%%%%%%%%%%%%%%%%
% BEGIN LAWN SWITCH POINT PLOT
% \begin{figure*}[htbp!]
% \centering
% %\subfigure[\textcolor{red}{REMOVE ME}]{%
%     \pgfplotsset{every axis/.append style={
%         font=\large,
%         line width=2pt,
%         mark size=2.5pt,
%         tick style={line width=0.8pt}
%     }}
%     \begin{tikzpicture}[scale=0.8]
%     	\begin{semilogxaxis}[
% 		    xlabel=Epochs,ylabel=Test Accuracy,
% 		    ymin=93.5, ymax=94.5,
% 		    ymajorgrids,
% 		    xmajorgrids,
% 		    legend style={at={(0.03,0.03)},
% 		    anchor=south west}]
%         	\addplot[color=teal,mark=square*] coordinates {
%         		(1, 94.1)
%         		(5, 94.16)
%         		(20, 94.21)
%         		(40, 94.0)
%         	};
%         	\addplot[color=green, mark=square*] coordinates {
%         		(1, 94.25)
%         		(5, 94.03)
%         		(20, 94.14)
%         		(40, 94.06)
%         	};
%         	\addplot[color=olive, mark=square*] 
%         	coordinates {
%         		(1, 94.17)
%         		(5, 93.87)
%         		(20, 93.93)
%         		(40, 93.77)
%         	};
%         	legend style={at={(0.03,0.5)},anchor=west}
%         \legend{Batch size = 256, Batch size = 4k, Batch size = 10k}
%         \end{semilogxaxis}%
%     \end{tikzpicture}%
% %\label{fig:adam-cifar10-efree}}
% \caption{Effect of LAWN switch point $E_{free}$ on CIFAR-10 with Adam-LAWN}
% \label{fig:cifar-switching}
% \end{figure*}
% END LAWN SWITCH POINT PLOT
%%%%%%%%%%%%%%%%%%%%%%%%%%%
%%%%%%%%%%%%%%%%%%%%%%%%%%%

\subsection{Recommendation Systems Experimental Setup}

\textbf{Dataset Details}
Table~\ref{tab:recodatasets} contains statistics about the 3 datasets used.

%%% TABLE FOR DATASETS FOR RECO SYSTEMS
\begin{table}[h!]

  \begin{center}

    \begin{tabular}{l|c|c|r}
      \toprule % <-- Toprule here
      \textbf{Dataset} & \textbf{n(samples)} & \textbf{n(users)} & \textbf{n(items)}\\
      \midrule % <-- Midrule here
      MovieLens-100k & $\sim$100k & 943 & 1682\\
      MovieLens-1M & $\sim$1M & 6040 & 3706\\
      Pinterest & $\sim$1.5M & 54906 & 9909\\
      \bottomrule % <-- Bottomrule here
    \end{tabular}
  \end{center}
  \caption{Datasets for item recommendation.}
  \label{tab:recodatasets}
\end{table}

\textbf{Dataset Pre-Processing} Following the experimental setup used in He et al~\cite{he2017neural}, we first remove all users in the 3 datasets that have less than 20 interactions. We then binarize the data by treating any interactions as 1s (positives) and the absence of an interaction as 0s (negatives). While training, we sample four negatives for every positive. The negatives are refreshed every epoch. 

\textbf{Network Architecture} We adopt the multi-layer perceptron (MLP) architecture used by He et al~\cite{he2017neural}. The MLP is used in conjunction with user and item embeddings. The embeddings are concatenated and pushed through the MLP. The MLP follows a tower pattern, and results in a scalar output prediction $y_{ui}$ for a user $u$ and item $i$, which can then be compared to the ground truth label $y$ using the binary cross entropy loss for training and evaluation purposes. The number of units in each layer of the MLP are 256-128-64-1, where the first layer takes a 256-dimensional vector as input. Consequently, user and item embeddings are of 128 dimensions each.

\textbf{Evaluation Details} For each user in the datasets, we consider the latest interaction with an item for test performance evaluation. Specifically, we use the \textit{hit ratio@10} metric, where the test item is part of 100 randomly sampled items that the user has not interacted with in the past. The metrics records whether the test item occurs in the top 10 of the predicted ranking of the 100 items. 

\textbf{Hyperparameters}

We tuned the following parameters: learning rate, $E_{free}$ and weight decay. Momentum for SGD and SGD-LAWN was fixed at 0.9. $E_{total}$ was fixed at 300 epochs for all experiments, and $E_{warmup}$ was fixed at 30 epochs.

For LAWN optimizers, we tuned $E_{free}$ and did not use any weight decay pre- or post-switch to constrained training. For regular optimizers, we tuned weight decay. We used a different grid of learning rates for SGD and SGD-LAWN, in comparison to other optimizers. This is because SGD usually requires a higher learning rate than adaptive optimizers for optimal performance. 

Table~\ref{tab:reco-parameters} contains information about the different hyperparameters that were tuned. Results with standard error can be found in Tables~\ref{tab:res_se_ml100k}, \ref{tab:res_se_ml1m} and \ref{tab:res_se_pin}. Figure~\ref{fig:adam-ml100k-bs} shows the effect of $E_{free}$ on MovieLens-100k when Adam-LAWN is used.

\begin{table}[h]
\centering
\begin{tabular}{ll}
\hline
\textbf{Parameter} & \textbf{Values}                                             \\ \hline
$E_{warmup}$                          & 30                                             \\
$E_{free}$ (LAWN only)                & [0.1, 1, 5, 20, 50, 100]                        \\
Weight decay (non-LAWN only)     & [1e-4, 1e-3, 1e-2, 1e-1, 0]                      \\
Learning rate (non-SGD)                   & [1e-5, 5e-5, 1e-4, 5e-4, 1e-3, 5e-3, 1e-2, 5e-2] \\
Learning rate (SGD and SGD-LAWN)                   & [1e-2, 5e-2, 1e-1, 0.25, 0.5, 0.75, 1.0, 2.0, 2.5, 3.75, 5.0] \\ \hline
                                 &                                                 
\end{tabular}
\caption{Parameter grid for LAWN and regular optimizers.}
\label{tab:reco-parameters}
\end{table}

%%%%%%%%%%%%%%%%%%%%%%%%%%%
%%%%%%%%%%%%%%%%%%%%%%%%%%%
% BEGIN MAIN TABLE FOR RECO SYSTEMS
\begin{table*}[!ht]
\centering
% \resizebox{0.5\textwidth}{!}{%
\begin{tabular}{l|llll}
\hline
\multirow{2}{*}{\textbf{Method}}  &
  \multicolumn{3}{c}{\textbf{MovieLens-100k}} \Tstrut\\
  %\hline
    &
  \multicolumn{1}{c}{1k} &
  \multicolumn{1}{c}{10k} &
  \multicolumn{1}{c}{100k}  &
  \multicolumn{1}{c}{400k}  
  \Bstrut\\
  \hline
SGD    & 66.33 (0.24)&  65.58 (0.17)&  Fail & Fail \Tstrut\\
SGD-L  & \textbf{66.91} (0.24)&  \textbf{66.49} (0.16)&  Fail & Fail \Bstrut\\
\hline
Adam   & 66.01 (0.25) & 66.03 (0.15) & 63.20 (0.24) & 63.98 (0.25)  \Tstrut\\
Adam-L & \textbf{66.81} (0.17) & \textbf{66.91} (0.15) & \textbf{66.24} (0.16) & \textbf{66.14} (0.22)  \Bstrut\\
\hline
LAMB   & 65.45 (0.25) & 65.34 (0.16) & 64.23 (0.16) & 62.57 (0.17)  \Tstrut\\
LAMB-L & \textbf{66.56} (0.22) & \textbf{66.54} (0.20) & \textbf{66.52} (0.25) & \textbf{66.14} (0.23)  \Bstrut\\
\hline
\end{tabular}
%}
\caption{Test HR@10 on MovieLens-100k with standard error.}
\label{tab:res_se_ml100k}
\end{table*}

\begin{table*}[!ht]
\centering
% \resizebox{0.5\textwidth}{!}{%
\begin{tabular}{l|llll}
\hline
\multirow{2}{*}{\textbf{Method}}  &
  \multicolumn{3}{c}{\textbf{MovieLens-1M}} \Tstrut\\
  %\hline
    &
  \multicolumn{1}{c}{1k} &
  \multicolumn{1}{c}{10k} &
  \multicolumn{1}{c}{100k}  &
  \multicolumn{1}{c}{1M}  
  \Bstrut\\
  \hline
SGD    & 70.91 (0.18)&  69.31 (0.17)&  Fail & Fail \Tstrut\\
SGD-L  & 70.91 (0.12)&  \textbf{70.41} (0.05)&  Fail & Fail \Bstrut\\
\hline
Adam   & 69.87 (0.07) & 70.12 (0.11) & 69.28 (0.10) & 68.99 (0.21)  \Tstrut\\
Adam-L & \textbf{70.80} (0.09) & \textbf{70.41} (0.20) & \textbf{70.77} (0.04) & \textbf{70.66} (0.22)  \Bstrut\\
\hline
LAMB   & 69.91 (0.12) & 69.77 (0.09) & 69.44 (0.10) & 68.95 (0.12)  \Tstrut\\
LAMB-L & \textbf{70.86} (0.21) & \textbf{70.86} (0.04) & \textbf{70.68} (0.08) & \textbf{70.34} (0.19)  \Bstrut\\
\hline
\end{tabular}
%}
\caption{Test HR@10 on MovieLens-1M with standard error.}
\label{tab:res_se_ml1m}
\end{table*}

\begin{table*}[!ht]
\centering
% \resizebox{0.5\textwidth}{!}{%
\begin{tabular}{l|llll}
\hline
\multirow{2}{*}{\textbf{Method}}  &
  \multicolumn{3}{c}{\textbf{Pinterest}} \Tstrut\\
  %\hline
    &
  \multicolumn{1}{c}{1k} &
  \multicolumn{1}{c}{10k} &
  \multicolumn{1}{c}{100k}  &
  \multicolumn{1}{c}{1M}  
  \Bstrut\\
  \hline
SGD    & \textbf{86.62} (0.05)&  85.57 (0.20)&  Fail & Fail \Tstrut\\
SGD-L  & 86.61 (0.20)&  \textbf{85.99} (0.20)&  Fail & Fail \Bstrut\\
\hline
Adam   & \textbf{87.27} (0.07) & 85.97 (0.01) & 85.81 (0.01) & 85.30 (0.20)  \Tstrut\\
Adam-L & 86.85 (0.10) & \textbf{86.61} (0.10) & \textbf{86.04} (0.10) & \textbf{86.06} (0.20)  \Bstrut\\
\hline
LAMB   & 86.63 (0.10) & 85.91 (0.10) & 85.80 (0.10) & 85.65 (0.20)  \Tstrut\\
LAMB-L & \textbf{86.83} (0.10) & \textbf{86.25} (0.10) & \textbf{85.99} (0.10) & \textbf{86.07} (0.10)  \Bstrut\\
\hline
\end{tabular}
%}
\caption{Test HR@10 on Pinterest with standard error.}
\label{tab:res_se_pin}
\end{table*}

% END MAIN TABLE FOR RECO SYSTEMS
%%%%%%%%%%%%%%%%%%%%%%%%%%%
%%%%%%%%%%%%%%%%%%%%%%%%%%%

\subsection{ImageNet Experimental Setup}

\textbf{Dataset Details} The ImageNet dataset consists of about 1.2 million training images and 50,000 validation images. 

\subsubsection{Data Pre-processing} 
This model uses the following data augmentation:

For training:
\begin{itemize}
    \item Normalization (0 mean, 1 std. dev.)
    \item Random resized crop to $224 \times 224$
    \item Scale from $8\%$ to $100\%$
    \item Aspect ratio from $3/4$ to $4/3$
    \item Random horizontal flip
\end{itemize}

For inference:
\begin{itemize}
    \item Normalization (0 mean, 1 std. dev.)
    \item Scale to $256 \times 256$
    \item Center crop to $224 \times 224$
\end{itemize}

\subsubsection{Model}

We used the ResNet-50-v1.5 model for ImageNet classification, which differs slightly from the original ResNet50 model\cite{he2015deep}. For more details, the reader can refer to this open source code repository \url{https://github.com/NVIDIA/DeepLearningExamples}.

\subsubsection{The LAMB+ algorithm}
The LAMB algorithm uses a ratio consisting of a transformation of the network weight norm in the numerator and the update (including contribution of weight decay) in the denominator. This is known as the \textbf{trust ratio}. Without any or small weight decay, the trust ratio can grow to a large value and cause the training to become unstable. To counter this, we cap the trust ratio to have a max value of 1.0 (Algorithm \ref{algo:lamb}). We found this to make training more stable, and to have better generalization performance. We refer to this algorithm as LAMB+.

\subsubsection{Hyperparameter tuning}

All optimizers were tuned with a fixed epoch budget of 90 epochs regardless of batch size. For optimizers that use weight decay, we did not apply weight decay to batch normalization parameters. 

Momentum for SGD and SGD-LAWN was fixed at 0.875. Gradient clipping was used for all optimizers except SGD and SGD-LAWN, with the gradient norm clipped to a maximum value of 1.0.

\textbf{Small batch size}

 For SGD at a  batch size of 256, we grid searched the learning rate over the set $\{0.1, 0.256, 0.5\}$ and fixed weight decay to $10^{-4}$. Following \cite{goyal2018accurate}, we allow 0.5 epochs for $E_{warmup}$ and then decay the learning rate linearly to zero at the end of training. For SGD-LAWN at a  batch size of 256, we retained the peak learning rate from the best SGD setting and tuned $E_{free}$ and $E_{warmup}$ over a small grid of $\{1, 5, 10\}$. We used weight decay during the $E_{free}$ phase to arrest the growth of network weights, but not during the LAWN phase.

For Adam, we grid searched the learning rate over the set $\{0.0001, 0.00125, 0.0025, 0.005\}$ and did not use any L2 regularization or weight decay. We used 0.16 epochs for $E_{warmup}$ and used linear decay. For Adam-LAWN, we grid searched the learning rate from $\{0.0003, 0.0006, 0.0008, 0.004\}$. Similar to SGD-LAWN, we tuned $E_{free}$ and $E_{warmup}$ over a small grid of $\{1, 5, 10\}$.

For LAMB and LAMB+, we bootstrapped using suggestions in literature and fixed weight decay to 0.01 and $E_{warmup}$ to 0.16 epochs. We grid searched the learning rate over $\{0.001, 0.002, 0.0035, 0.005\}$. For LAMB-LAWN, we fixed weight decay to 0.01 for the pre-LAWN phase and $E_{free}$ to match learning rate warmup of LAMB to 0.16 epochs. We tuned $E_{warmup}$ over a grid of $\{1, 5, 10\}$ and used the same grid for learning rate as LAMB.

% and learning over a grid of $\{0.001, 0.002, 0.0035, 0.005\}$. 

\textbf{Large batch size}

For SGD for large batch size, we retain the weight decay value to be $10^{-4}$. Since learning rate warmup is important for large batch sizes \cite{goyal2018accurate}, we tune $E_{warmup}$ over a grid of $\{0.16, 1, 5, 10\}$ epochs. We also tuned the learning rate over the values $\{1.0, 5.0, 10.0\}$. For SGD-LAWN for large batch size, we used the same grid for learning rate as SGD. We tuned $E_{free}$ and $E_{warmup}$ over a small grid of $\{1, 5, 10\}$.

For Adam at a batch size of 16k, we tuned peak learning rate over a grid of $\{0.001, 0.003, 0.005, 0.008, 0.01\}$, weight decay over $\{0, 1e-3, 1e-2\}$ and duration of learning rate warmup over $\{5, 10, 20\}$. For Adam-LAWN for large batch size, we used the same grid for learning rate as Adam. Similar to SGD-LAWN and Adam-LAWN at small batch sizes, we tuned $E_{free}$ and $E_{warmup}$ over a small grid of $\{1, 5, 10\}$.

For LAMB and LAMB+, we again fixed weight decay to 0.01. We grid searched the learning rate over $\{0.01, 0.02, 0.02828, 0.03, 0.04\}$. We chose $0.02828$ because that's the reported learning rate for LAMB on ImageNet for batch size of 16k. We grid searched $E_{warmup}$ over $\{5, 10, 20\}$.

For LAMB-LAWN, we retained the best settings of learning rate and weight decay from LAMB and tuned $E_{free}$ and $E_{warmup}$ over a small grid of $\{5, 10, 20\}$.

\subsection{LAWN vs. Other methods for controlling loss of adaptivity}
We now provide details of the experiments for table~\ref{tab:lawn-vs-other} where we compare LAWN to other methods for controlling loss of adaptivity. 

The dataset was fixed to MovieLens-1M, and experiments were run for batch sizes of 10k and 100k. Learning rate was tuned using a small grid of $\{0.001, 0.005, 0.01\}$ for each method. For flooding, we tuned the flooding parameter over the grid $\{0.2, 0.3, 0.4\}$. For LSR, the $\alpha$ parameter was fixed at 0.05. All experiments were run 3 times, and the average of the runs was reported.

\section{Max margin formulations - Fully homogeneous, Over-parameterized nets}
\label{sec:marginproof}

In this section, we establish equivalence between normalized margin maximization (see (\ref{eq:normmarginmax})) and its constrained form (see (\ref{eq:constrmaxmargin})), to which LAWN is related. Then, using similar arguments, we also explain how $\ell_2$ regularization is related to normalized margin maximization.

%\anika{Intro sentence as to why LAWN and L2 separate (like how this is organized, purpose of this appendix)?}

{\bf LAWN.} We first prove that, for fully homogeneous nets, the norm constrained maximization of margin is equivalent to normalized margin maximization.
While Banburski et al~\cite{Banburski2019} proves a similar result, its statement and proof are imprecise, and not in the form needed to support LAWN. Hence, we provide a new formulation and a more detailed demonstration.
%Banburski et al~\cite{Banburski2019} proves a similar result, but its statement and proof are imprecise. Also, they are not in the precise form needed to show the connection with LAWN. Hence we give full details here.

Let $\phi_i(w)$ denote the target class logit for example $i$, where $w=\{w^{\ell}\}$ is the weight vector of the fully homogeneous network with $w^{\ell}$ being the weight subvector of layer $\ell$. Let $\{c^{\ell}\}$ be any set of positive constants.

{\bf Lemma 1.} The following optimization formulations are equivalent:
\begin{equation}
 \max_{\{v^{\ell}\}} \frac{\min_i \phi_i (\{v^{\ell}\})}{\Pi_{\ell} \|v^{\ell}\|} \label{eq:normmarginform}
\end{equation}
\begin{equation}
 \max_{\{w^{\ell}\}} \min_i \phi_i (\{w^{\ell}\}) \mbox{ s.t. } \|w^{\ell}\| = c^{\ell} \;\; \forall \ell
 \label{eq:lawnform}
\end{equation}
and, at their respective (local) optima, for each $\ell$, $v^{\ell}$ and $w^{\ell}$ are along the same direction.\footnote{Since the $\phi_i$ are homogeneous in each $v^{\ell}$, the objective function in (\ref{eq:normmarginform}) is unaffected by the scale (one individual scale for each $\ell$) of the $v^{\ell}$ and hence (\ref{eq:normmarginform}) has a continuum of solutions (a conic manifold of dimension equal to the number of $\ell$'s). A solution, $\{w^\ell\}$ of (\ref{eq:lawnform}) is one specific normalized solution on that manifold.}

{\bf Proof.} Let us start from (\ref{eq:lawnform}) and reduce it to (\ref{eq:normmarginform}). We begin by rewriting (\ref{eq:lawnform}) as
\begin{eqnarray}
 \max_{\gamma\in R_+, \{w^{\ell}\}} \gamma  \;\; \mbox{ s.t. } \;\; \|w^{\ell}\| = c^{\ell} \;\; \forall \ell, \;\; \phi_i (\{w^{\ell}\}) \ge \gamma \; \forall i
 \label{eq:lawnform2}
\end{eqnarray}
For each $\ell$, let us reparameterize $w^{\ell}$ using a new set of variables $u^{\ell}$ as
\begin{equation}
    w^{\ell} = \frac{c^{\ell}}{\|u^{\ell}\|} u^{\ell} \label{eq:wvrelation}
\end{equation}
%\anika{Does this assume vl and wl are along same direction, or is not and this remains something we still need to prove?}
where each $u^{\ell}$ is now unconstrained. With this reparameterization, the equality constraints, $\|w^{\ell}\| = c^{\ell} \; \forall \ell$ in (\ref{eq:lawnform2}) are automatically satisfied; hence, they can be removed. This leads us to the equivalent formulation
\begin{eqnarray}
 \max_{\gamma\in R_+, \{u^{\ell}\}} \gamma  \;\; \mbox{ s.t. } \;\;  \phi_i (\{\frac{c^{\ell}}{\|u^{\ell}\|} u^{\ell}\}) \ge \gamma \; \forall i
 \label{eq:lawnform3}
\end{eqnarray}
By the homogeneity of each $\phi_i$ we get
\begin{eqnarray}
 \max_{\gamma\in R_+, \{u^{\ell}\}} \gamma  \;\; \mbox{ s.t. } \;\;  \phi_i (\{ u^{\ell}\}) \ge \gamma^\prime \; \forall i
 \label{eq:lawnform4}
\end{eqnarray}
where
\begin{equation}
    \gamma^\prime = \frac{\gamma \Pi_{\ell} \|u^{\ell}\|}{c}, \;\; \mbox{ and } \;\; c = \Pi_{\ell} c^{\ell}
\end{equation}
Using $\gamma^\prime$ instead of $\gamma$ as a new variable and omitting $c$ as a constant, we get the new problem
\begin{eqnarray}
 \max_{\gamma^\prime\in R_+, \{u^{\ell}\}} \frac{\gamma^\prime}{\Pi_{\ell} \|u^{\ell}\|}   \;\; \mbox{ s.t. } \;\;  \phi_i (\{ u^{\ell}\}) \ge \gamma^\prime \; \forall i
 \label{eq:lawnform5}
\end{eqnarray}
Now we can remove $\gamma^\prime$ as a variable and rewrite the problem as
\begin{eqnarray}
  \max_{\{u^{\ell}\}} \frac{\min_i \phi_i (\{u^{\ell}\})}{\Pi_{\ell} \|u^{\ell}\|}
 \label{eq:lawnform6}
\end{eqnarray}
which is the same as (\ref{eq:normmarginform}) with $u^{\ell}$ and $v^{\ell}$ being one and the same set of variables. This proves Lemma 1.

{\bf Remark 1.} The optimization problems in (\ref{eq:normmarginform}) and (\ref{eq:lawnform}) have local optima. Though not stated in any previous work, it is worth pointing out that, in general position, the optima of (\ref{eq:lawnform}) are isolated and finite in number. Transversality theorem of differential topology may be used to establish this. (In a single layer network the optimum is unique.) By appropriately defining and restricting to local neighborhoods of a local optimum, the equivalence indicated in Lemma 1 does apply to the local optima of (\ref{eq:normmarginform}) and (\ref{eq:lawnform}) too.

{\bf $\ell_2$ Regularization.} Lyu and Li~\cite{lyu2020gradient} shows that, instead of (\ref{eq:normmarginform}), implicit bias can also be expressed via the optimization problem,
\begin{equation}
 \max_{v} \frac{\min_i \phi_i (v)}{\|v\|^{\#\ell}} \label{eq:normmarginL2}
\end{equation}
where $\#\ell$ is the number of layers in the fully homogeneous network.
Then, using the same arguments used in the proof of Lemma~1, we can show that (\ref{eq:normmarginL2}) is equivalent to
\begin{equation}
 \max_{w} \min_i \phi_i (w) \;\; \mbox{ s.t. } \;\; \|w\| = c
 \label{eq:normmarginL2c}
\end{equation}
Like in~\S\ref{sec:method} (see the discussion around~(\ref{eq:appnormmarginmax})), we can approximate the margin, $\min_i \phi_i (w)$ by $h(L(w))$, use the monotonicity of $h$, and replace $\|w\|=c$ by $\|w\|^2=c^2$ to get the approximate problem,
\begin{equation}
    \min_w L(w) \;\; \mbox{ s.t. } \;\; \|w\|^2 = c^2
 \label{eq:normmarginL2c2}
\end{equation}
The $\ell_2$ regularization formulation,
\begin{eqnarray}
    L_{\ell_2} = L(w) + \frac{\rho}{2} \|w\|^2
    \label{eq:L22}
\end{eqnarray}
is nothing but the regularized equivalent of (\ref{eq:normmarginL2c2}), i.e., the trajectory of solutions of (\ref{eq:L22}) as $\rho$ goes from $0$ to $\infty$ is same as the trajectory of solutions of (\ref{eq:normmarginL2c2}) as $c$ goes from $\infty$ to $0$. Thus, like LAWN, $L_2$ regularization also approximates implicit bias, with the approximation becoming better as $\rho\rightarrow 0$.

\section{Explaining escape}
\label{sec:escape}

Recent theoretical and empirical analysis have pointed out that deep net training needs two key phases to yield good generalization~\citep{JastrzebskiSFAT20, Wu_NEURIPS2018} --
%\anika{change punctuation to --} 
{\em Phase 1:} Escaping poor minima; and {\em Phase 2:} Achieve implicit bias in the region of the chosen minimum. These phases are influenced by the training paradigm (the type of formulation, regularization, normalizations employed, etc.) and the training hyperparameters ($\eta$, the learning rate, and $B$, the batch size are the two crucial ones). 

Phase 1 of the training identifies which minimum's region of attraction that the training finally falls into. Escaping regions of attraction of "sharper" minima with sub-optimal generalization performance is an important function of phase 1. Two matrices - $\Sigma$, the covariance of the noise associated with minibatch gradient, and $H$, the Hessian of the training loss - combine with $\eta$ and $B$ to dictate what happens in phase 1. To give a rough guidance, Wu et al~\citep{Wu_NEURIPS2018} conduct a rough theoretical analysis around a minimum and require that the following condition is met for escape from that minimum:
\begin{equation}
\lambda_{\max} \left\{ (I-\eta H)^2 + \frac{\eta^2(m-B)}{B(m-1)} \Sigma \right \} > 1 \label{eq:Wucondn}
\end{equation}
where $\lambda_{\max}(A)$ denotes the largest eigenvalue of $A$, $m$ is the total number of training examples and, $\Sigma$ and $H$ are evaluated at the minimum. The following are useful to note. 
\begin{itemize}
\item As we move to large batch sizes and hence get $B$ closer to $m$, the effect of the second term diminishes significantly, causing the escaping to become difficult. This is a key reason why normal training methods suffer from a loss of generalization performance with large batch sizes.
\item When the target class logit values become very large and hence training loss starts moving very close to zero, $H$ and $\Sigma$ themselves become very small, thus severely hampering the escape. A careful scheduling of $\eta$ to very large values to cause the escape and then come back to smaller $\eta$ values to continue the training to a better minimum are needed. But designing such a schedule is hard.
\end{itemize}

Phase 2 of the training is where the optimization method implicitly affects the generalization performance; for example, several gradient based methods~\citep{lyu2020gradient, Poggio2019, Poggio2020, Wang2020, soudry2018implicit} lead to a direction that locally maximizes the normalized margin. 
%In normal, unregularized training, in this phase, weights and logits become large and training loss goes to zero. Unless large $\eta$ values are employed~~\citep{lyu2020gradient}, the implicit bias is not fully realized.

LAWN helps in both of the phases. First, by keeping weights (and hence logits) well bounded and keeping most examples away from regions where loss flattens to zero, $\Sigma$ and $H$ are kept large and hence escape from sub-optimal minima is made easier. In fact, even with large batch sizes where $B$ comes close to $m$, the largeness of $H$ can make escape possible. This is why, as demonstrated in~\S\ref{sec:experiments}, LAWN does much better than normal training with large batch sizes. Second, as argued in~\S\ref{sec:method}, the implicit bias (normalized margin maximization) properties are well-approximated by the constrained formulations used by LAWN, possibly helping performance to be improved in phase 2 too.

\end{document}